# Dimensionality reduction to maximize prediction generalization capability

Takuya Isomura[1,2*], Taro Toyoizumi[1,3*]

[1] Laboratory for Neural Computation and Adaptation, RIKEN Center for Brain Science, Wako, Saitama 351-0198, Japan

[2] Brain Intelligence Theory Unit, RIKEN Center for Brain Science, Wako, Saitama 351-0198, Japan

[3] Department of Mathematical Informatics, Graduate School of Information Science and Technology, The University of Tokyo, Bunkyo-ku, Tokyo 113-8656, Japan

[*] Corresponding authors email: takuya.isomura@riken.jp, taro.toyoizumi@riken.jp

**Abstract**

Generalization of time series prediction remains an important open issue in machine learning, wherein earlier methods have either large generalization error or local minima. We develop an analytically solvable, unsupervised learning scheme that extracts the most informative components for predicting future inputs, termed predictive principal component analysis (PredPCA). Our scheme can effectively remove unpredictable noise and minimize test prediction error through convex optimization. Mathematical analyses demonstrate that, provided with sufficient training samples and sufficiently high-dimensional observations, PredPCA can asymptotically identify hidden states, system parameters, and dimensionalities of canonical nonlinear generative processes, with a global convergence guarantee. We demonstrate the performance of PredPCA using sequential visual inputs comprising hand-digits, rotating 3D objects, and natural scenes. It reliably estimates distinct hidden states and predicts future outcomes of previously unseen test input data, based exclusively on noisy observations. The simple architecture and low computational cost of PredPCA are highly desirable for neuromorphic hardware.

Prediction is essential for both biological organisms [1–3] and machine learning [4–6]. In particular, they both need to predict the dynamics of newly encountered sensory inputs (i.e., test data) based on and only on knowledge learned from a limited number of past experiences (i.e., training data). Generalization error is a standard measure of the generalization capability of predicting the future consequences of previously unseen input data, which is defined as the difference between the training and test prediction errors. It is thus crucial for organisms and machines to find a prediction strategy with a small generalization error, as otherwise their predictions will fail because of overfitting to the training data.

Despite the importance of generalizing prediction, current mainstream machine learning approaches have some limitations. They can be categorized into three major groups, and their limitations are summarized as follows: (1) The most basic prediction strategy is to learn a direct mapping from past to future inputs in the form of an autoregressive (AR) model (**Fig. 1a**). Although AR models are simple to construct and guarantee global convergence, their predictions contain a large generalization error because the mapping from the observations to the prediction is often redundant, leading to severe overfitting when the number of training samples is limited [7,8]. Thus, to make accurate predictions, low-dimensional (i.e., concise) representations should be extracted from high-dimensional (i.e., redundant) sensory data. (2) A dimensionality reduction technique can be used to obtain a concise representation [9]; however, this is often achieved separately from the prediction step—e.g., by first applying an autoencoder to reduce the dimensionality [10,11] and then employing a long short-term memory to predict the sequence [12] (**Fig. 1b**). The first autoencoding step—which provides a low-dimensional representation that minimizes the loss for reconstructing the current input—is the most basic dimensionality reduction strategy. One problem of this approach is that autoencoders may preferentially extract observation noise that is useless for prediction, owing to its extra variance. From a prediction perspective, it is more helpful to reduce the dimensionality to minimize the prediction error, similar to the approach used in time-lagged autoencoders (TAEs) [13] and their variants [14,15] (**Fig. 1c**). These approaches combine predictions with dimensionality reduction in a single architecture. (3) A major approach to time-series prediction is to construct a state-space model (SSM). SSMs, which include the Kalman filter [16] and its nonlinear variants [17,18], simultaneously perform dimensionality reduction and prediction (**Fig. 1d**). From this model-based perspective, the best prediction is achieved when an SSM employs the states and parameters that match the true system properties. However, the problem becomes difficult when both the hidden states and system parameters are unknown. In particular, their predictions become inaccurate owing to nonlinear interactions between the uncertainties in hidden states and parameters, as they can create spurious solutions. Furthermore, the dimensionality of hidden states, which is essential for prediction accuracy, is difficult to optimize. Conventional model selection approaches using some information criterion [19–21], structural risk [22], or cross-validation [23] would fail to identify the optimal dimensionality when the state or parameter estimation converges to a suboptimal solution. In short, all three approaches have essential drawbacks that interfere with the generalization of accurate predictions.

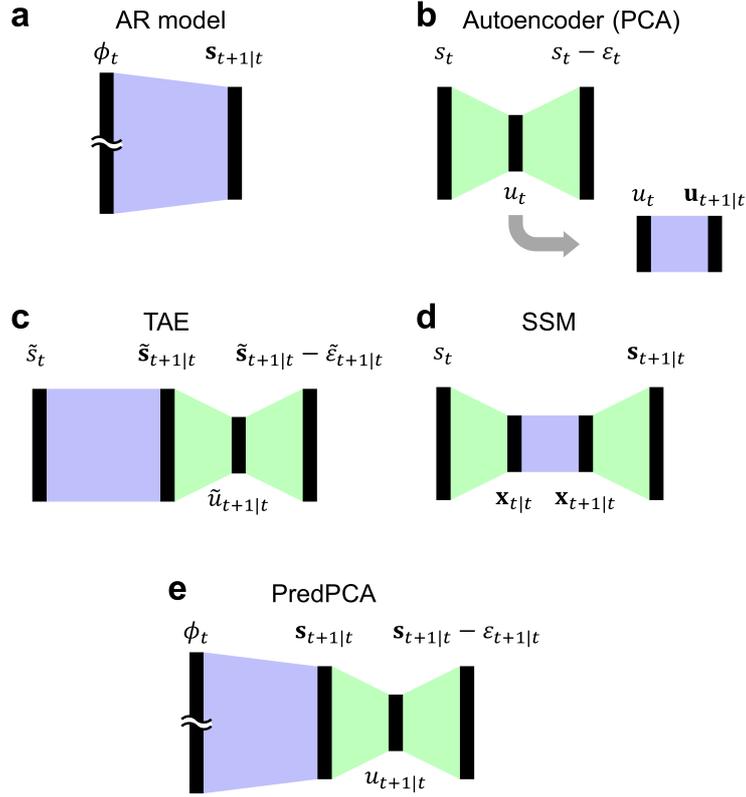

**Figure 1.** Five different prediction model structures. A black bar denotes a layer of a neural network, while blue and green trapezoids denote synaptic weight matrices for prediction and dimensionality reduction, respectively. (**a**) Naïve AR models directly compute the maximum likelihood estimator of the next input $\mathbf{s}_{t+1|t}$ based on the bases $\phi_t \equiv \left(s_t^T, s_{t-1}^T, \ldots, s_{t-K_p+1}^T\right)^T$ that summarize current and past observations. (**b**) Two-step prediction models first extract a concise representation $u_t$ using an autoencoder (PCA) by minimizing the loss $\varepsilon_t$ and then predict the next representation $\mathbf{u}_{t+1|t}$ using a recurrent neural network. (**c**) TAEs and their variants combine predictions with dimensionality reduction by performing eigenvalue decomposition or singular value decomposition of the transition matrix of the input. Note that $\tilde{s}_t \equiv \Sigma_s^{-1/2} s_t$ denotes the whitened inputs, $\tilde{\mathbf{s}}_{t+1|t} \equiv \Sigma_s^{-1/2} \mathbf{s}_{t+1|t}$ denotes the whitened predicted inputs, and $\Sigma_s$ denotes the actual input covariance. (**d**) SSMs update the hidden state estimator

$\mathbf{x}_{t|t}$ based on the previous state and current input, and predict the next state $\mathbf{x}_{t+1|t}$ and input $\mathbf{s}_{t+1|t}$. (**e**) PredPCA first computes the maximum likelihood estimator $\mathbf{s}_{t+1|t}$ based on multi-timestep basis functions $\phi_t$ and then extracts a concise representation $u_{t+1|t}$, by minimizing the prediction error $\varepsilon_{t+1|t}$. This scheme can effectively filter out the causes of the generalization error.

To overcome these limitations, we establish a method that can solve this simultaneous optimization problem of hidden states, system parameters, and dimensionality with a global convergence guarantee. We develop an unsupervised learning scheme for extracting features that are essential for prediction, termed *predictive principal component analysis* (PredPCA). It is formally derived from the minimization of the squared prediction error and can extract low-dimensional predictive features from high-dimensional sensory inputs, even in the presence of observation noise that is much larger than the signals themselves. This robustness is because PredPCA conducts *post-hoc dimensionality reduction* to extract a concise representation of the predicted input (**Fig. 1e**), unlike autoencoders or SSMs. Moreover, the architecture of PredPCA is suitable for noise reduction because it predicts the subsequent input based on multi-timestep basis functions, unlike TAEs and their variants. These properties allow PredPCA to find hidden states (c.f., blind source separation [24]) and perform long-term prediction reliably and accurately. In particular, system parameter identification [25,26] using PredPCA contrasts with conventional methods. It is guaranteed to asymptotically identify the true parameters of canonical nonlinear systems (see below for the definition) in the large sample-size limit, when the mappings from hidden states to sensory inputs are sufficiently high-dimensional. In addition, based on Akaike's statistics [19,27], we analytically derive a mathematical formula that estimates the test prediction error of PredPCA. It shows that the generalization error is proportional to an entropy due to the sampling fluctuation [27]. The minimization of this formula can optimize unknown free parameters, including the rank of system dimensions and number of past observations used for prediction, and can provide the global minimum of the test prediction error. We mathematically and numerically demonstrate that filtering out unpredictable noise by using PredPCA is essential to maximizing the prediction generalization capability.

**RESULTS**

**Predictive principal components analysis (PredPCA)**

In this work, we assume that hidden states $x_t$ generate higher-dimensional sensory inputs $s_t$ as follows:

$$s_t = g(x_t) + \omega_t, \tag{1}$$

and the dynamics of hidden states are described by

$$x_{t+1} = f(x_t, x_{t-1}, x_{t-2}, \ldots) + z_t, \tag{2}$$

where $z_t$ and $\omega_t$ are mutually independent white noises, with zero means and covariances $\Sigma_z$

and $\Sigma_\omega$ (**Fig. 2a**, left, and Methods A). Process noise $z_t$ adds stochasticity into the hidden state dynamics, while observation noise $\omega_t$ represents any unpredictable fluctuations that we wish to remove. PredPCA is applicable to sensory data generated from systems involving either Gaussian or non-Gaussian noise to extract features and characterize system properties, although the identification of noise distributions is more straightforward when provided with Gaussian noise (see Methods F). Although this paper focuses on white noise, PredPCA's outcomes would also be accurate with colored noise as long as the auto-correlation time constant of $\omega_t$ is smaller than that of $x_t$. To apply PredPCA to continuous-time systems, the time bin size should be determined depending on their applications. **Table 1** presents the glossary of expressions.

PredPCA aims to extract the components containing the most information for predicting the next input $s_{t+1}$ based on current and past observations $s_t, s_{t-1}, \ldots, s_{t-K_p+1}$. With this in mind, we consider a linear neural network whose output is given by

$$u_{t+1|t} = V\phi_t, \tag{3}$$

where $u_{t+1|t}$ is an $N_u$-dimensional vector of encoders, $V$ is a (horizontally long) $N_u \times N_\phi$ encoding synaptic weight matrix, and $\phi_t \equiv \left(s_t^T, s_{t-1}^T, \ldots, s_{t-K_p+1}^T\right)^T$ is an $N_\phi$-dimensional vector of linear basis functions that summarize current and past observations. We refer to this linear encoder as a neural network intending an analogy to biological neural networks and to highlight potential applications to neuromorphic computation (see Discussion for further details). Unlike standard principal component analysis (PCA) [28,29] and autoencoders [10,11], which minimize the reconstruction error in the current input, PredPCA minimizes the prediction error $\varepsilon_{t+1|t} \equiv s_{t+1} - W^T u_{t+1|t}$, defined as the difference between the actual next input at *t+1* and the prediction based on inputs up to *t*. Here, $W^T$ is an $N_s \times N_u$ decoding synaptic weight matrix used for predicting the next input $s_{t+1}$ based on the concise encoders $u_{t+1|t}$ (where we introduced $W^T$ rather than $W$ for a notational reason that will become clear below). PredPCA's cost function $L$ is defined as the expectation of the squared prediction error over the training period *T*:

$$L \equiv \frac{1}{2} \left\langle |\varepsilon_{t+1|t}|^2 \right\rangle_q. \quad (4)$$

Here, $\langle \cdot \rangle_q \equiv \frac{1}{T} \sum_{t=1}^{T} \cdot$ indicates the expectation over the empirical distribution *q*. By minimizing this cost function with respect to $V$, we obtain the optimal encoding weights as $V = W\mathbf{Q}$, where $\mathbf{Q} \equiv \langle s_{t+1} \phi_t^T \rangle_q \langle \phi_t \phi_t^T \rangle_q^{-1}$ (Methods B). Thus, $u_{t+1|t} = W \mathbf{s}_{t+1|t}$ holds, where $\mathbf{s}_{t+1|t} = \mathbf{Q}\phi_t$ is the maximum likelihood estimator of $s_{t+1}$. The synaptic weight matrix $W$ is updated by gradient descent on $L$. After some additional transformations (Methods B), we obtain

$$\dot{W} \propto -\frac{\partial L}{\partial W} = \left\langle u_{t+1|t} \left( s_{t+1} - W^T u_{t+1|t} \right)^T \right\rangle_q. \quad (5)$$

The fixed point of equation (5) yields the transpose of optimal decoding weights that minimize $L$. The solution ensures that the encoders $u_{t+1|t}$ achieve the optimal representation for prediction.

Equation (5) is equivalent to the subspace rule of PCA [28], except that $u_{t+1|t}$ encodes the future state at time *t+1* instead of the state at time *t* (i.e., the standard PCA uses $u_{t|t}$). This means that PredPCA, which is defined by the prediction error minimization, can be decomposed into two steps: computing the maximum likelihood estimator of $s_{t+1}$, $\mathbf{s}_{t+1|t}$, followed by a post-hoc PCA of $s_{t+1|t}$ using the eigenvalue decomposition (**Fig. 1e**). Owing to the global convergence property of the subspace rule for PCA [30], the global convergence of equation (5) is also guaranteed. In essence, PredPCA is a convex optimization. Crucially, however, only PredPCA (but not the standard PCA) can effectively filter out unpredictable observation noise, as we demonstrate numerically below and mathematically in Methods C. It is straightforward to extend PredPCA to multi-step predictions (see Methods B for further details). Note that although this paper focuses on the prediction of subsequent inputs (i.e., autoregression), it is straightforward to apply PredPCA to minimize the generalization error for a class of regression tasks. (The formulation for this is performed simply by supposing that the hidden states $x_t$ generate both observations $s_t$ and a high-dimensional target signal $y_t$ and by replacing the prediction error $\varepsilon_{t+1|t}$ with $\varepsilon_t \equiv y_t - W^T V \phi_t$.)

After extracting the hidden states by using PredPCA, we employ independent component analysis (ICA) [31,32], which can separate the extracted states into independent components as long as the true hidden states of the external milieu are actually mutually independent. For example, when the network observes a sequence of MNIST handwritten digits [33], PredPCA followed by ICA generates 10-dimensional independent encoders $\mathbf{x}_{t+1|t}$, each element of which encodes one of the 10 possible digits (**Fig. 2a**, right). The detailed procedure to extract $\mathbf{x}_{t+1|t}$ from $u_{t+1|t}$ is provided in Methods E.

Previous works have developed methods combining future data predictions with dimensionality reduction. In particular, one may be reminded of time-lagged independent component analysis (TICA) [14], TAE [13], and dynamic mode decomposition (DMD) [15]. When $\phi_t = s_t$, PredPCA is involved in this family of methods—thus, one may view PredPCA as a combination of these methods and AR models based on high-dimensional, multi-timestep basis functions. This construction enables PredPCA to effectively filter out observation noise and reduce test prediction error (see below).

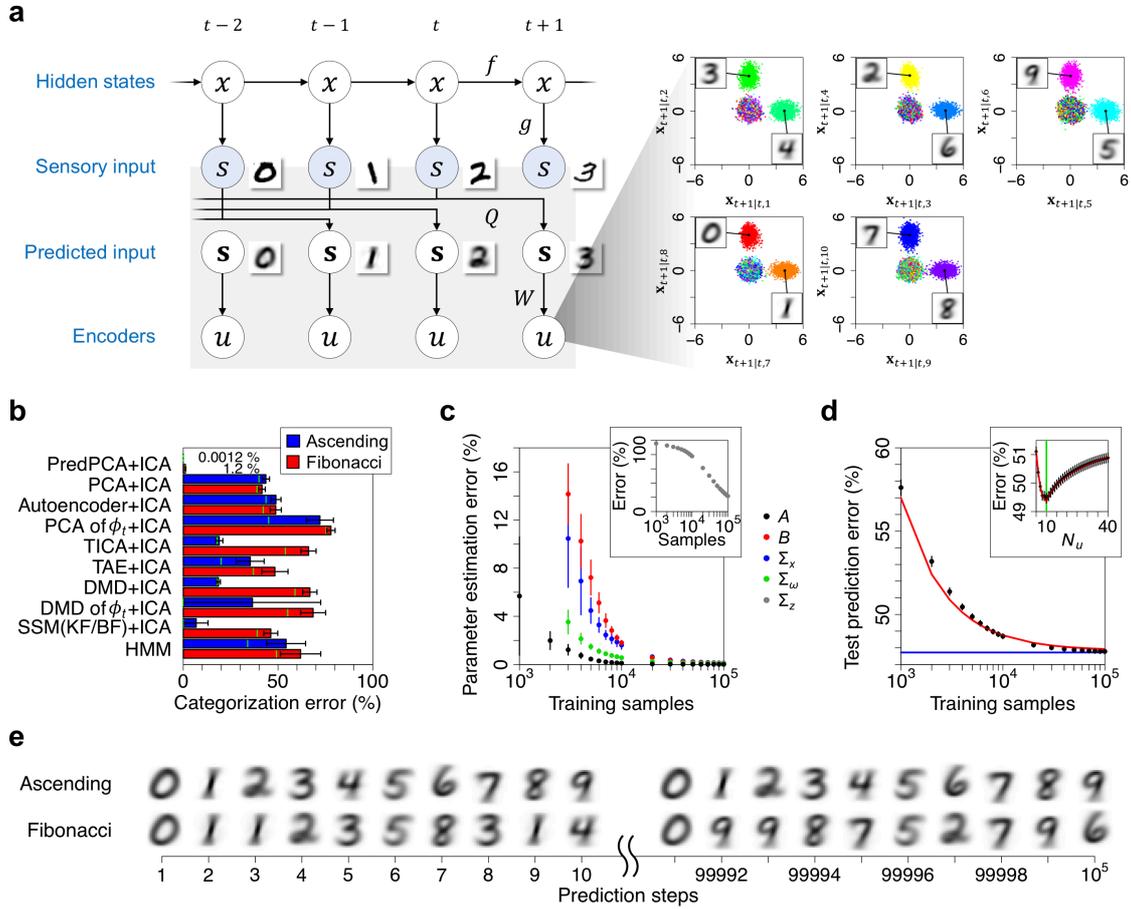

**Figure 2.** PredPCA of handwritten digit sequences. (**a**) Left: System comprising a generative process (top) and a neural network that follows PredPCA (bottom, shaded). The network is trained with an image sequence $s_t$ of handwritten digits generated from the dynamics of 10-dimensional hidden states $x_t$, each element of which expresses one of the 10 digits. Right: 10-dimensional

independent encoders (i.e., hidden state estimator) $\mathbf{x}_{t+1|t}$ obtained using PredPCA and ICA. $2\times10^4$ test samples that are color-coded by their digit are plotted. (**b**) Comparison with related methods in terms of the mean categorization error (i.e., false discovery rate), obtained by averaging categorization errors of 10 elements of $\mathbf{x}_{t+1|t}$. The digits are introduced in ascending order (blue) and Fibonacci sequence (red). An SSM based on a Kalman filter (KF) is used for ascending sequence, while that based on a Bayesian filter (BF) is used for the Fibonacci sequence. The green bars indicate the minimum categorization error among 20 different realizations of digit sequences. The error bars indicate the standard deviation. (**c**) Parameter estimation error measured by the squared Frobenius norm ratio, where the difference between the ground truth parameter matrix $\theta$ and its estimator $\mathbf{\theta}$ is divided by their norm, $error = |\mathbf{\theta} - \theta|_F^2 / \max(|\theta|_F^2, |\mathbf{\theta}|_F^2)$. We assume here that the ascending-order handwritten digit sequence is generated from a linear system comprising $s_t = Ax_t + \omega_t$ and $x_{t+1} = Bx_t + z_t$ ($A$, black, $B$, red). The covariance matrices $\Sigma_x$ (blue), $\Sigma_\omega$ (green), and $\Sigma_z$ (gray, inset) are associated with $x_t, \omega_t, z_t$, respectively. (**d**) Test error in predicting the next handwritten digit images in the ascending-order sequence, measured by the normalized mean squared error over test samples, $error = \langle |s_{t+1} - W^T u_{t+1|t}|^2 \rangle / \langle |s_{t+1}|^2 \rangle$. The red line (in the main and inset panels) represents a theoretical prediction obtained using equation (7). The blue line denotes the lower bound of the error, calculated via supervised learning. The inset panel depicts the dependence of the test prediction error on the encoding dimensionality $N_u$ (when $T = 6000$), where $N_u = 10$ (green line) is optimal. (b)(c)(d) are obtained with 20 different realizations of digit sequences. Note that the error bars in (d) are hidden by the circles. (**e**) Long-term prediction using PredPCA and ICA. A winner-takes-all operation is applied to make greedy predictions of the digit sequences. After receiving the first 40 digits, unless those initial digit images are outliers, the network can predict the next $10^5$ digits (and more) without any categorization error. See Supplementary Methods 1 and 2 for further details.

**Key analytical discoveries**

We conducted comprehensive mathematical analyses to rigorously demonstrate the performance and statistical properties of PredPCA. In particular, we demonstrated the following two key properties: (1) It is mathematically guaranteed that PredPCA can identify the *optimal* (explained below) hidden state representation and parameter estimators—up to a linear transformation that does not affect prediction accuracy—for general linear systems and, asymptotically, even for nonlinear systems (Methods E, F). When equations (1) and (2) are involved in a class of canonical nonlinear systems defined by the undermentioned equations (8) and (9), a set of hidden states, parameters, and dimensionalities that characterize a system is uniquely determined up to a trivial linear ambiguity (Methods A). Under this condition, while using a linear neural network for the encoding, the asymptotic linearization theorem [34] ensures that PredPCA will extract the true hidden states when the hidden state dimensionality is large and the input dimensionality is sufficiently larger than the hidden state dimensionality. Briefly, this is because projecting the high-dimensional input onto the directions of the major eigenvectors of the input covariance effectively magnifies the linearly transformed components of the hidden states included in the input, while filtering out the nonlinear components (see Methods E for its mathematical statement and the conditions for application; see [34] for the mathematical proof).

Owing to this linearization property, the hidden state estimator $\mathbf{x}_{t+1|t}$ obtained using PredPCA asymptotically converges to a linear transformation of the maximum likelihood estimator of hidden states $x_{t+1}$, i.e., $\langle x_{t+1}\phi_t^T\rangle_q\langle\phi_t\phi_t^T\rangle_q^{-1}\phi_t$. Hence, PredPCA provides the optimal hidden state representation for prediction. Furthermore, the analytical expressions of the system parameter estimators are derived as functions of $\mathbf{x}_{t+1|t}$, with a convergence guarantee to the true parameter values in the large sample-size and system-size limits. These parameter estimators are calculated by a simple iteration-free computation summarized in **Table 2** and Methods F. In essence, provided with sufficient but finite training samples, PredPCA can identify the hidden states and parameters of large-scale canonical systems up to a small estimation error. This result is surprising because the reliable identification of the optimal hidden states and the true parameters were previously only described within the framework of supervised learning, whereas PredPCA can provide them by unsupervised learning without relying on the true hidden states $x_t$.

(2) PredPCA can maximize the prediction generalization capability by minimizing the test prediction error

$$L_{test} \equiv \frac{1}{2}\left\langle\left|\varepsilon_{t+1|t}\right|^2\right\rangle. \tag{6}$$

Here, $\langle\cdot\rangle \equiv \int \cdot\, p(\phi_t, s_{t+1})d\phi_t ds_{t+1}$ indicates the expectation over the true distribution

$p(\phi_t, s_{t+1})$ (note the difference from equation (4)). In practice, however, the true distribution is unknown for a learner. Thus, one needs to estimate equation (6) based on and only on parameters estimated based on training data. In the framework of the maximum likelihood estimation or squared error minimization, the expectation of the test error is expressed as an Akaike information criterion (AIC) [19] or network information criterion (NIC) [20], respectively. Similar to the derivation of AIC and NIC, we explicitly compute the expectation of equation (6), with the optimized synaptic weights, as

$$\underbrace{\mathcal{L}}_{\substack{\text{test error} \\ \text{expectation}}} \equiv \mathrm{E}_{\{q\}}[L_{test}] = \underbrace{\frac{1}{2}\left(\mathrm{tr}[\mathbf{\Sigma}_s] - \mathrm{tr}[\mathbf{P}_s^T \mathbf{\Sigma}_s^{\text{Pred}} \mathbf{P}_s]\right)}_{\text{training error}} + \underbrace{\frac{N_\phi}{2T} \mathrm{tr}[\mathbf{P}_s^T (\mathbf{\Sigma}_s - \mathbf{\Sigma}_s^{\text{Pred}}) \mathbf{P}_s]}_{\text{generalization error}} + \mathcal{O}\left(T^{-\frac{3}{2}}\right). \quad (7)$$

Methods D presents the derivation. Here, $T$ is the number of training samples, $\mathbf{P}_s$ is the first-to-$N_u$-th major eigenvectors of the predicted input covariance $\mathbf{\Sigma}_s^{\text{Pred}} \equiv \langle \mathbf{s}_{t+1|t} \mathbf{s}_{t+1|t}^T \rangle_q$ (where $W^T W = \mathbf{P}_s \mathbf{P}_s^T$ holds at the fixed point of equation (5)), and $\mathbf{\Sigma}_s \equiv \langle s_t s_t^T \rangle_q$ is the actual input covariance. The expectation $\mathrm{E}_{\{q\}}[\cdot]$ is taken over different empirical distributions $q$, each of which comprises $T$ training samples and is used to optimize synaptic weights.

The expectation of the test prediction error $\mathcal{L}$ is characterized by two free parameters: the rank of encoding dimensions ($N_u$) and the number of past observations used for the maximum likelihood estimation ($K_p$), where $N_\phi = K_p N_s$. Optimizations of $N_u$ and $N_\phi$ performed while updating the synaptic weight using equation (5) provide the global minimum of $\mathcal{L}$. The optimal encoding dimensionality is guaranteed to converge to the true hidden basis dimensionality of the canonical system for a large but finite $T$ (Methods D). The second term of $\mathcal{L}$, referred to as the generalization error, is associated with an entropy due to the sampling fluctuation [27]. This term indicates that only the prediction error projected to the major eigenspace causes the generalization error, which highlights the importance of dimensionality reduction to reduce the test prediction error. In short, naïvely minimizing the training error by using a large encoding dimensionality, such as in AR models, leads to overfitting; in contrast, minimizing $\mathcal{L}$ provides the best encoding dimensionality and number of past observations to generalize the prediction.

For further details, please see Methods and Supplementary Information. The aforementioned analytical results are empirically validated though numerical simulations by confirming the reliable identification of properties of canonical systems defined in Methods A (**Supplementary Fig. 1a–c**). Furthermore, empirical observations imply that the outcomes of PredPCA can be utilized to identify the properties of more general classes of systems (e.g., a class involving a Lorenz attractor; **Supplementary Fig. 1d–f**), although system parameter identification beyond the class defined in Methods A has not yet been proven mathematically. In what follows, we demonstrate the performance of PredPCA using sequential visual inputs comprising handwritten digits, rotating 3D objects, and natural scenes (refer to Supplementary Methods 1 and 2 for simulation protocols).

**PredPCA provides optimal representation and parameters for prediction**

In the first experiment (**Fig. 2**), we trained a neural network with MNIST handwritten digit images [33] in ascending order and in the Fibonacci sequence, wherein only the last digit was presented; however, these sequences involve some additional stochasticity (which corresponds to process noise $z_t$) such that a digit was replaced by a random one and a monochrome inversion occurred with a small probability at each step (as an analog to large noise that interferes with weak signal measurements: e.g., movement artifacts in electroencephalogram recordings [35]). In both cases, PredPCA successfully extracted 10-dimensional features underlying the image sequences as they were relevant to predicting the sequences. The following ICA [32] separated the extracted components into independent hidden states. Each of the ensuing encoder neurons (i.e., independent components, $\mathbf{x}_{t+1|t}$) selectively responded to one of the 10 digits without being taught their labels, as we can see for the encoders trained with the ascending sequence in **Fig. 2a** (right).

  Irrespective of the sequence types (ascending order and Fibonacci sequence), PredPCA and ICA precisely separated the digits into 10 clusters in 10 dimensions with an average categorization error of less than 2% (scored by false discovery rate; **Fig. 2b**). During this process, PredPCA ignored any within-class differences in the digit images that do not predict the next image (which correspond to observation noise $\omega_t$). Hence, PredPCA's policy of dimensionality reduction to minimize the prediction error distinguishes it from standard PCA [28,29] and autoencoders [10,11]—because PCA and autoencoders minimize the reconstruction error for the current input $s_t$ and thus preferentially extract the within-class differences in the digit images due to their extra variances. Even when the standard PCA was applied to the past-to-current input sequence (i.e., $\phi_t$), it failed to separate the digits because the hidden representation of $\phi_t$ included more than 10-dimensional state space and thus the first 10 major components of $\phi_t$ did not match the true hidden states $x_t$. Although the categorization errors of TICA [14], TAE [13], and DMD [15] were smaller than those of PCA and an autoencoder, they still failed to categorize some digits. This is because the former methods use only a single step (i.e., $\phi_t = s_t$) to predict subsequent digit images ($s_{t+1}$). Moreover, even when using $(s_t, s_{t-1})$ or $\phi_t$ to predict $(s_{t+1}, s_t)$ or $\phi_{t+1}$, they failed to categorize digits, as the extracted features do not match the true hidden states (these results are similar to the PCA of $\phi_t$). The performance of the SSM and hidden Markov model (HMM) with 10-dimensional state spaces was also poor because their larger parameter estimation errors led them to a spurious solution (i.e., local minimum).

In addition to accurate source separation, PredPCA could provide the optimal system parameters for the prediction (**Fig. 2c**). These parameter estimators were computed simply by following the definitions in **Table 2**. The differences between the parameter estimators obtained by PredPCA and those obtained by supervised learning converged to zero as the number of training samples increased, as predicted theoretically (Methods F). These results validated that PredPCA-based system parameter identification was applicable to systems involving non-Gaussian noise. Consequently, the outcomes of PredPCA could reliably identify the transition rules underlying the ascending order (**Extended Data Fig. 1a**) and Fibonacci sequences (**Extended Data Fig. 1b**) in an unsupervised manner. In essence, we demonstrated that each encoder obtained using PredPCA corresponds to a digit, and the obtained state transition matrix represents the estimated dynamics of digit sequences, which can assign the meaning to these model parameters.

These results indicate the interpretability of PredPCA as the obtained model can provide an explanation of the manner that the hidden dynamics generate the sensory input.

The above outcomes allowed PredPCA to predict subsequent digits reliably and accurately (**Fig. 2d**). Here, we see that although PredPCA did not observe the hidden states directly, its test prediction error converged globally—with increasing training samples—to the lower bound of the test prediction error computed via supervised learning that explicitly used the true hidden states for training. This is as theoretically predicted by equation (7). Moreover, equation (7) successfully identified the optimal encoding dimensionality that minimized the test prediction error as $N_u = 10$, which also matched the true hidden state dimensionality (**Fig. 2d**, inset panel). These matchings hold even in the absence of random replacement and/or monochrome inversion of digit images (**Extended Data Fig. 1c**). Numerical observations indicate that PredPCA can reduce errors in categorization, system identification, and prediction as the number of past observations used for prediction ($K_p$) increases until reaching its finite optimum (**Extended Data Fig. 1d**). In contrast, linear TAE and SSM (same as PredPCA with $\phi_t = s_t$) failed to identify the system properties, and thus generated a larger prediction error (**Extended Data Fig. 12**).

In particular, the long-term prediction of subsequent digits highlights the significance of PredPCA's categorization and system identification accuracy—provided with a winner-takes-all operation, the outcomes of PredPCA could recursively predict the subsequent digits without categorization errors for more than $10^5$ steps (**Fig. 2e**). These results were minimally influenced by the assumed transition mapping structures and training history (**Extended Data Fig. 3a–c**), and the optimal model structure could be determined through model selection based on the standard AIC (**Extended Data Fig. 3d**). In contrast, SSMs tended to fail the long-term prediction depending on initial conditions and training history, even though provided with the winner-takes-all operation (**Extended Data Fig. 3e**).

**PredPCA filters out observation noise and minimizes test prediction error**

Next, the noise reduction and prediction generalization capabilities of PredPCA were examined using natural movies. We trained a neural network by using images of 3D objects rotating anti-clockwise [36] as the input (**Fig. 3a**, furthest left). In short, the task was to predict the opposite side of test object images (200 objects) by observing only a half side of the images, based on the transition (i.e., rotational) mapping learned from different training object images (up to 800 objects). Here, we used the optimal linear bases $\phi_t$ to maximize PredPCA's generalization capability (see Supplementary Methods S3 for the procedure). The significance of PredPCA was experimentally confirmed by its successful predictions of the 30°–150° rotated images of previously unseen test objects (**Fig. 3a**, middle row; see **Supplementary Movie 1** for predictions of 90° rotated images, where the right-hand-side images are the predictions of the corresponding ground truth images on the left-hand side).

In general, features extracted using PredPCA comprise categorical features that represent *what* the input is as well as dynamical features that express *how* it is moving. As the asymptotic linearization theorem implies that the obtained encoders $\mathbf{x}_{t+k|t}$ are linear superpositions of hidden states, we define categorical features as the average of estimators, $\bar{\mathbf{x}}_t \equiv (\mathbf{x}_{t+30|t} + \cdots + \mathbf{x}_{t+150|t})/5$, and dynamical features as the deviation from their average, $\Delta \mathbf{x}_{t+k|t} \equiv \mathbf{x}_{t+k|t} - \bar{\mathbf{x}}_t$. Applying ICA to $\bar{\mathbf{x}}_t$ separated the categorical features into a sparse representation, each dimension of which expresses a feature of objects (**Fig. 3b**, top). Applying an additional PCA to the dynamical features (e.g., $\Delta \mathbf{x}_{t+90|t}$) provided the angle of 3D objects as the first principal component (PC1; **Fig. 3c**, left). Although the coordinate of the attractor changes depending on its category, this treatment makes it easier to interpret the dynamics of hidden states. We also observed the same property for $\mathbf{x}_{t+30|t}$, $\mathbf{x}_{t+60|t}$, $\mathbf{x}_{t+120|t}$, and $\mathbf{x}_{t+150|t}$.

Notably, these prediction and feature extraction capabilities were largely retained even in the presence of an artificially added large (white Gaussian) observation noise whose variance had the same magnitude as the variance of original images, demonstrating the robustness of PredPCA's outcomes (**Fig. 3a,b**, bottom, and **Fig. 3c**, right; see **Supplementary Movie 2** for predictions of 90° rotated images). The sampling fluctuation caused by the observation noise disturbed the prediction of minor components, and thus changed the optimal encoding dimensionality (**Fig. 3d**).

We confirmed an earlier decrease of the test prediction error for PredPCA relative to the naïve AR model as the number of training samples increases (**Fig. 3e**). PredPCA generated a smaller test prediction error relative to TICA, TAE, DMD, and SSMs based on Kalman or Bayesian filters (**Fig. 3f**). These results indicate that PredPCA could determine the plausible rule for rotating generic objects. Remarkably, owing to the convex optimization, features extracted using PredPCA are uniquely determined for any given training dataset (even if the true system is unknown). This contrasts with TAE and SSM, because their extracted features change depending on the initial conditions, order of supplying mini batches, or level of observation noise, even though they are trained with the same dataset (**Extended Data Fig. 4**).

Note that we also trained PredPCA with an image dataset of rotating 3D human faces and confirmed that PredPCA can accurately predict subsequent images and extract relevant features such as pan and tilt angles of face images in an unsupervised manner (**Supplementary Fig. 2**).

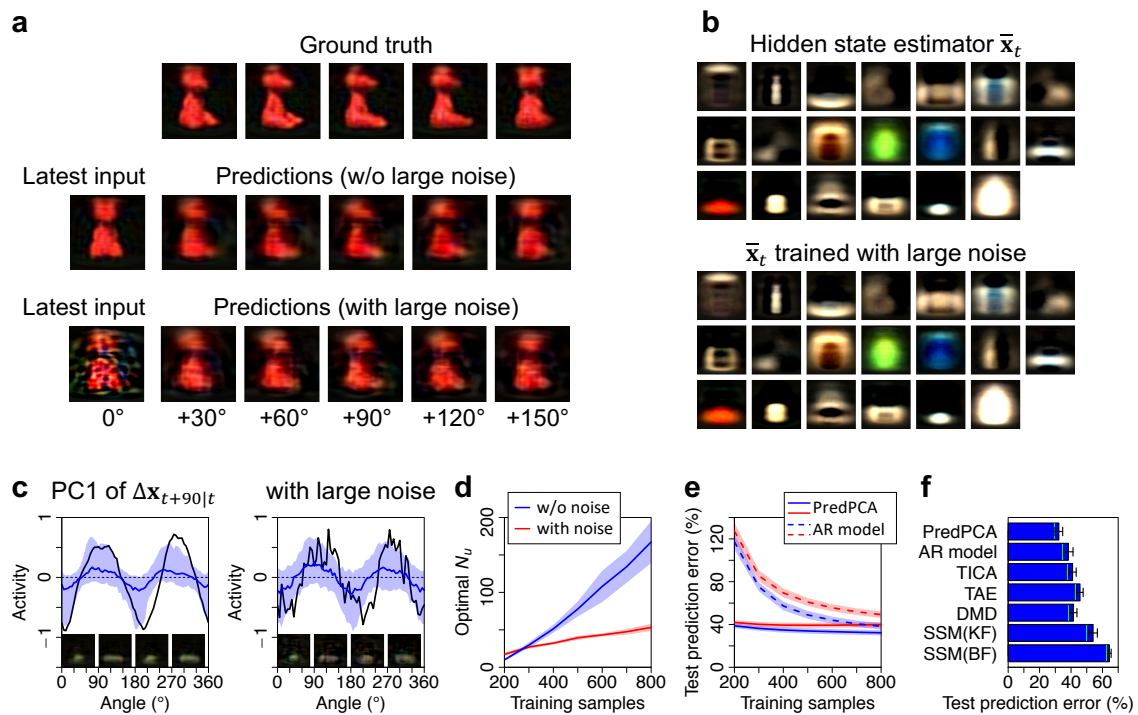

**Figure 3.** PredPCA-based denoising, hidden state extraction, and subsequent input prediction of movies of rotating 3D objects. (**a**) Snapshots of the prediction results. Latest input image (furthest left) and ground truth (top) and predicted images after 30°, 60°, 90°, 120°, and 150° rotations,

without (middle row) and with (bottom) artificially added observation noise. (**b**) Images corresponding to 20-dimensional sparse representations ($\bar{\mathbf{x}}_t$) each expressing a categorical feature of objects. These images were obtained by applying ICA with super-Gaussian prior distribution to the first 20 principal components of PredPCA, averaged over different prediction points $\bar{\mathbf{x}}_t = (\mathbf{x}_{t+30|t} + \cdots + \mathbf{x}_{t+150|t})/5$ (see Methods E for the detail). These images visualize linear mappings from each independent component to the observation. (**c**) Rotation of objects encoded in the first principal component of the dynamical features of $\mathbf{x}_{t+90|t}$ (i.e., $\Delta\mathbf{x}_{t+90|t} = \mathbf{x}_{t+90|t} - \bar{\mathbf{x}}_t$). Here, the neural activity predicted the angle of 90° rotated future images, indicating that when observing an asymmetric object (as opposed to a cylindrical object), the network was able to anticipate whether its image would be wider or narrower after a 90° rotation. Blue lines and shaded areas indicate the median and area between the 25th and 75th percentiles over the dataset, whereas black lines show trajectories for an object. (**d**) Optimal encoding dimensionality increasing with training sample size, in the absence (blue) and presence (red) of the large observation noise. (**e**) Comparison of test prediction error, defined by $error_k \equiv \langle |g_{t+k} - W^T u_{t+k|t}|^2 \rangle / \langle |g_{t+k}|^2 \rangle$, where $g_t \equiv s_t - \omega_t$ indicates the observation-noise-free input. PredPCA (solid lines) show a smaller test prediction error and an earlier error convergence compared with the naïve AR model (dashed lines). Blue and red lines denote the error in the absence and presence of the large observation noise. (**f**) Comparison of test prediction error between PredPCA, naïve AR model, TICA, TAE, DMD, and SSMs based on Kalman filter (KF) and Bayesian filter (BF), when trained with 800 objects in the absence of noise. (d)(e)(f) are obtained with 10 different realizations of training and test samples. The green bars in (f) indicate the minimum test prediction error among these 10 different realizations. The shaded areas and error bars indicate the standard deviation. See Supplementary Methods 1 and 2 for further details.

As a further application to more natural data, we lastly trained a neural network with natural scenes captured from a driving car [37] (**Fig. 4a** and **Supplementary Movie 3**). Here, we aimed at demonstrating the applicability of our simple, analytically solvable linear method to real-world video prediction and feature extraction tasks, rather than comparing the prediction accuracy of PredPCA with that of state-of-the-art video prediction methods exploiting engineering wisdom. For predictions, we separated the movies into six groups of data based on the magnitude of change in the images per frame and trained six predictors separately with each group of data; subsequently, post-hoc PCA was applied to the synthesized predicted input (see Supplementary Methods 1 for further details). PredPCA could predict 0.5 s future images of previously unexperienced natural scenes with a certain accuracy (**Fig. 4a**). Moreover, PredPCA could extract brightness, the vertical and lateral asymmetries, and lateral motion in sceneries underlying the driving car movies (**Fig. 4b,c**). For the feature extractions, the entire movie was simply supplied to PredPCA without the six sub-groups; thus, the global convergence was theoretically guaranteed. We observed tight correspondences between features learned based on different finite training samples (**Fig. 4b,c**, inset panels), implying that PredPCA could extract features unique to the generative process that generated sensory data. In particular, the PC1 of dynamical features that encoded the lateral motion in sceneries (**Fig. 4c**) was relevant to predicting the steering of the car. The features extracted using PredPCA were retained even when using the data grouping (**Extended Data Fig. 5a, b**). The other major categorical and dynamical features represent different categories of sceneries (**Extended Data Fig. 5c**) and motions in different positions (**Extended Data Fig. 5d**), respectively. Remarkably, unlike conventional video prediction methods [6], PredPCA could extract these relevant features in an unsupervised manner, without the use of labels or target signals for training.

In summary, using examples of 3D-rotating objects and natural scenes, we demonstrated that PredPCA can filter out observation noise and minimize test prediction error by extracting features relevant to generalizing predictions. Although the true generative process is unknown for these examples, these results indicate that the outcomes of PredPCA capture the plausible properties of natural data. These results highlight the prediction generalization and feature extraction capabilities of PredPCA as well as its wide applicability to real-world data.

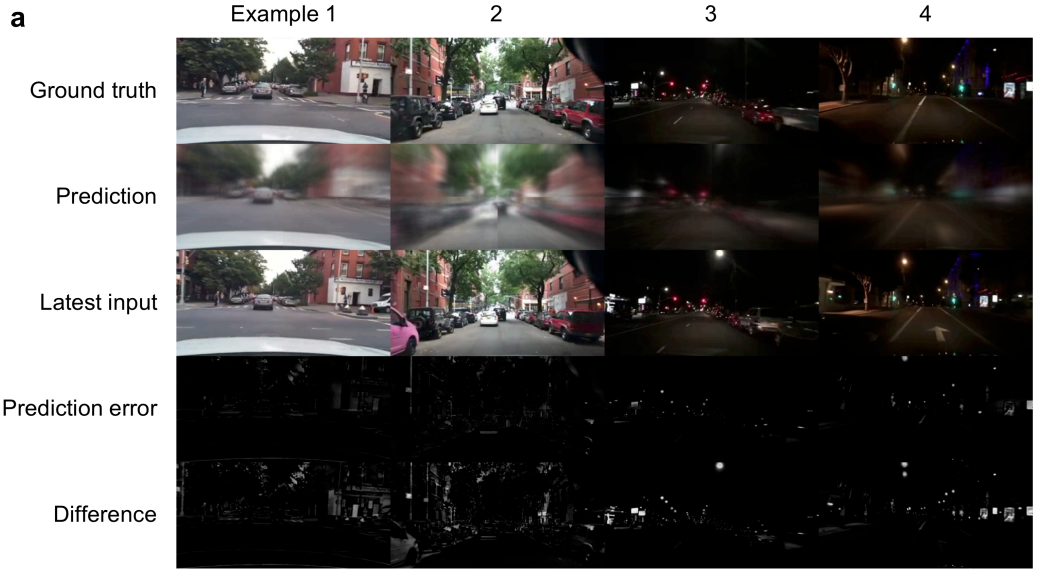
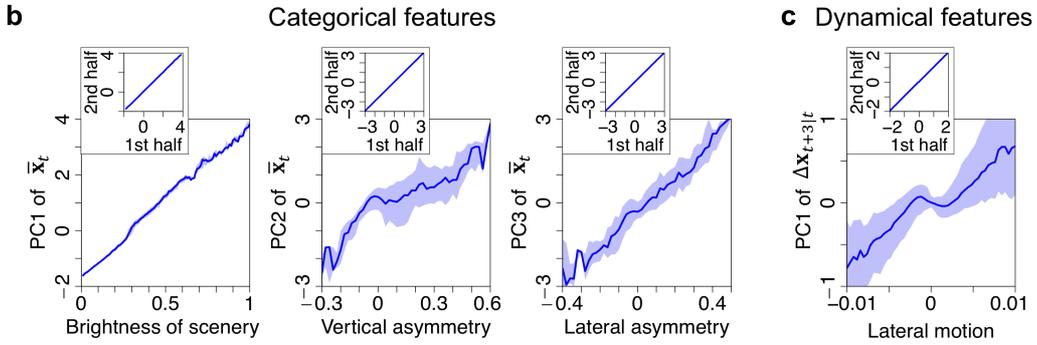

**Figure 4.** PredPCA of natural scene movies. (**a**) Four examples of predictions. Top: Ground truth or target image ($s_{t+15}$); i.e., 0.5 s future image of the latest input. Second row: Predicted image obtained using PredPCA ($W^T u_{t+15|t}$). It should be noted that the blurry edges in the predicted images occurred primarily because PredPCA predicted the mean of future outcome images. Third row: Latest input image ($s_t$). Fourth row: Prediction error between ground truth and predicted images; the white regions indicate the extent of errors (magnitude of $s_{t+15} - W^T u_{t+15|t}$). Bottom: Difference between ground truth and latest input images (magnitude of $s_{t+15} - s_t$). The test prediction error, defined by $error_{15} \equiv \langle |s_{t+15} - W^T u_{t+15|t}|^2 \rangle / \langle |s_{t+15} - s_t|^2 \rangle$, was 0.648. Although unexpected events during 0.5 s were unpredictable and some predictions were inaccurate due to the limited effective dimensionality of the input, the results indicate that PredPCA provides predictions that interpolate unseen future images and latest input images, without using any label for training. (**b**) Extraction of brightness and the vertical and lateral asymmetries in diving car movies as the PC1–PC3 of categorical features (i.e., $\bar{\mathbf{x}}_t$). Inset panels depict tight correspondences between features extracted using PredPCA, learned based exclusively on the first and second half of training samples. For the PC1-PC3, the correlations between them are larger than 0.9999. (**c**) Extraction of lateral motion from diving car movies as the PC1 of dynamical features (i.e., $\Delta\mathbf{x}_{t+3|t}$). The correlation between features learned based exclusively on the first and second half of training samples is 0.9962. In (b)(c), blue lines and shaded areas indicate the median and area between the 25th and 75th percentiles. Refer to Supplementary Methods 1 for further details.

**DISCUSSION**

Our proposed scheme, PredPCA, was shown to identify a concise representation that provides the global minimum of the test prediction error, by first predicting subsequent observations and then performing post-hoc PCA of the predicted inputs. This is essential for maximizing the prediction generalization capability, as well as for ensuring the accurate and unbiased estimation of system properties, comprising hidden states, system parameters, and dimensionalities. Our scheme is formally based on Akaike's statistics [19,27] and is consistent with existing information-theoretical views of biological optimizations, including maximum negentropy [38], predictive coding [2,3], predictive information [39], and the free energy principle [40]—providing them a normative, analytically-solvable example of a neural network that maximizes information quality and generalization capability.

PredPCA offers an interpretable hidden state representation (Methods E) preferable for generalizing prediction, without using prior knowledge about external systems. To this end, the global convergence guarantee or convex optimization of PredPCA (Methods B, D) is essential—because the representation could otherwise change depending on the initial conditions, training history, or level of observation noise, rendering such representation unreliable, and may overfit to a particular training data set. PredPCA further guarantees the asymptotic identification of the true system properties with a global convergence guarantee (Methods E, F and **Table 2**) when sensory data are generated from a class of canonical systems (Methods A), provided with sufficient training samples $T$ and sufficiently high-dimensional observations satisfying $N_s \gg N_x \gg 1$. This is remarkable because PredPCA can identify properties of canonical systems even with a limited number of training samples, up to a small range of errors that are inversely proportional to the sample and system sizes, as empirically validated in **Fig. 2** (and **Extended Data Fig. 1** and **Supplementary Fig. 1**; see also [34]). These guarantees are crucial, particularly when feature extraction failures or misunderstanding of a system lead to catastrophic problems in subsequent applications, such as in automated driving or medical diagnosis. In general, finite but sufficiently large $T$, $N_s/N_x$, and $N_x$ are required to ensure this asymptotic property—because only under such a condition, the major principal components of the de-noised input are guaranteed to match the hidden states of the original nonlinear system [34].

Unlike PredPCA, conventional (nonlinear) prediction strategies using autoencoders [10,11], TAE [13], or SSMs [16–18] do not have such guarantees and can fail to reliably provide accurate prediction and feature extraction depending on various conditions, as shown in **Figs. 2 and 3** and Methods C, because they have many spurious solutions. Having said this, although this paper primarily considers a different condition, if a learner has a sufficient amount of prior knowledge about the generative process that generates sensory data (e.g., knowledge about underlying physics), incorporating them into prediction can provide more interpretable and accurate predictions. Such knowledge may remove spurious solutions and render all solutions as the global minimum. In other words, for these related methods, the outcomes of PredPCA are potentially of great importance in setting a plausible initial condition and an appropriate empirical prior, in the absence of prior knowledge.

It should be noted that if the number of training samples is sufficient and the magnitude of observation noise is sufficiently low, the prediction error of PredPCA may be larger than that of state-of-the-art prediction methods using deep neural networks [4–6]—because the generalization error may be negligible under such a condition. The improvement in generalization capability obtained by omitting minor eigenmodes has been reported using deep neural networks [41,42]. This implies a potential extension of PredPCA to analytically solve the optimal representations for deep learning in a weakly nonlinear regime.

Although this work focuses on discrete-time systems, one may think of these systems as approximations of the physical reality in continuous time, where generative processes exhibit a hierarchy of timescales. From this perspective, the definitions of signal and observation noise will change depending on the time bin size of observations. Thus, it is crucial for accurate predictions to determine the time bin size to ensure that the timescale of observations matches the timescale of generative processes.

PredPCA-type learning can be implemented in biological neuronal networks and biologically inspired neuromorphic hardware [43,44]. Neurons in these systems must update their synapses to perform predictions under physiological or physical constraints—in particular, it is difficult for them to access non-local information, such as the synaptic weights of other neurons [45]. This fact limits biologically plausible learning to be a local rule that updates synapses based on and only on pre- and post-synaptic neural activities and additional directly accessible signals. As conventional PCA and ICA algorithms are non-local, we have developed a local learning algorithm that performs both PCA and ICA [46,47]. This algorithm can make PredPCA a local learning rule and guarantees its biological plausibility. Hence, one can speculate that PredPCA-type learning underlies the generalization capability of biological organisms and the self-organization of internal model [48]. Some neuronal substrate, such as neuromodulators [49,50], may encode the test prediction error expectation for mediating structural learning or model selection in the brain.

For further discussion, please refer to Supplementary Discussion.

In summary, PredPCA was proven to be an analytically solvable unsupervised dimensionality reduction scheme capable of extracting the most informative components for generalizing prediction. By effectively filtering out unpredictable noise, PredPCA can reliably identify plausible system properties, with a global convergence guarantee, and can globally minimize the test prediction error. Although this paper focuses on the autoregression, PredPCA can minimize the generalization error for a class of regression tasks, indicating its potential applicability to various real-world applications. As a mathematically proven optimal generalization strategy, our scheme is potentially useful for understanding biological generalization mechanisms and for creating reliable and explainable artificial general intelligence.

**Data availability**

Image data used in this work are available in the MNIST data set [33] (http://yann.lecun.com/exdb/mnist/index.html, for Fig. 2), the ALOI data set [36] (http://aloi.science.uva.nl, for Fig. 3), and the BDD100K data set [37] (https://bdd-data.berkeley.edu, for Fig. 4). Figures 2-4 are generated by applying our scripts (see below) to these image data.

**Code availability**

MATLAB scripts used in this work are available at https://github.com/takuyaisomura/predpca or https://doi.org/10.5281/zenodo.4362249. The scripts are covered under the GNU General Public License v3.0.

**References**

1. Rao, R. P. & Ballard, D. H. Predictive coding in the visual cortex: a functional interpretation of some extra-classical receptive-field effects. *Nat. Neurosci.* **2**, 79-87 (1999).

2. Rao, R. P. & Sejnowski, T. J. Predictive sequence learning in recurrent neocortical circuits. *Adv. Neural Info. Proc. Syst.* **12**, 164-170 (2000).

3. Friston, K. A theory of cortical responses. *Philos. Trans. R. Soc. Lond. B Biol. Sci.* **360**, 815-836 (2005).

4. Srivastava, N., Mansimov, E. & Salakhudinov, R. Unsupervised learning of video representations using lstms. *Int. Conf. Machine Learning* 843-852 (2015).

5. Mathieu, M., Couprie, C. & LeCun, Y. Deep multi-scale video prediction beyond mean square error. Preprint at https://arxiv.org/abs/1511.05440 (2015).

6. Lotter, W., Kreiman, G. & Cox, D. Deep predictive coding networks for video prediction and unsupervised learning. Preprint at https://arxiv.org/abs/1605.08104 (2016).

7. Hurvich, C. M. & Tsai, C. L. Regression and time series model selection in small samples. *Biometrika* **76**, 297-307 (1989).

8. Hurvich, C. M. & Tsai, C. L. A corrected Akaike information criterion for vector autoregressive model selection. *J. Time Series Anal.* **14**, 271-279 (1993).

9. Cunningham, J. P. & Ghahramani, Z. Linear dimensionality reduction: Survey, insights, and generalizations. *J. Mach. Learn. Res.* **16**, 2859-2900 (2015).

10. Hinton, G. E. & Salakhutdinov, R. R. Reducing the dimensionality of data with neural networks. *Science* **313**, 504-507 (2006).

11. Kingma, D. P. & Welling, M. Auto-encoding variational bayes. Preprint at https://arxiv.org/abs/1312.6114 (2013).

12. Hochreiter, S. & Schmidhuber, J. Long short-term memory. *Neural Comput.* **9**, 1735-1780 (1997).

13. Wehmeyer, C. & Noé, F. Time-lagged autoencoders: Deep learning of slow collective variables for molecular kinetics. *J. Chem. Phys.* **148**, 241703 (2018).

14. Pérez-Hernández, G., Paul, F., Giorgino, T., De Fabritiis, G. & Noé, F. Identification of slow molecular order parameters for Markov model construction. *J. Chem. Phys.* **139**, 015102 (2013).

15. Klus, S., Nüske, F., Koltai, P., Wu, H., Kevrekidis, I., Schütte, C. & Noé, F. Data-driven model reduction and transfer operator approximation. *J. Nonlinear Sci.* **28**, 985-1010 (2018).

16. Kalman, R. E. A new approach to linear filtering and prediction problems. *J. Basic Eng.* **82**, 35-45 (1960).

17. Julier, S. J. & Uhlmann, J. K. New extension of the Kalman filter to nonlinear systems. In *Signal processing, sensor fusion, and target recognition VI* (Vol. 3068, pp. 182-193). (International Society for Optics and Photonics, 1997).

18. Friston, K. J., Trujillo-Barreto, N. & Daunizeau, J. DEM: A variational treatment of dynamic systems. *NeuroImage* **41**, 849-885 (2008).

19. Akaike, H. A new look at the statistical model identification. *IEEE Trans. Automat. Contr.* **19**, 716-723 (1974).

20. Murata, N., Yoshizawa, S. & Amari, S. I. Network information criterion-determining the number of hidden units for an artificial neural network model. *IEEE Trans. Neural Netw.* **5**, 865-872 (1994).

21. Schwarz, G. Estimating the dimension of a model. *Ann. Stat.* **6**, 461-464 (1978).

22. Vapnik, V. Principles of risk minimization for learning theory. *Adv. Neural Info. Proc. Syst.* **4**, 831-838 (1992).

23. Arlot, S. & Celisse, A. A survey of cross-validation procedures for model selection. *Stat. Surv.* **4**, 40-79 (2010).

24. Comon, P. & Jutten, C. (Eds.). Handbook of Blind Source Separation: Independent component analysis and applications. (Academic press, 2010).

25. Ljung, L. System Identification: Theory for the User (2nd ed.). (Prentice-Hall, 1999).

26. Schoukens, J. & Ljung, L. Nonlinear System Identification: A User-Oriented Roadmap. Preprint at https://arxiv.org/abs/1902.00683 (2019).

27. Akaike, H. Prediction and entropy. In *Selected Papers of Hirotugu Akaike* (pp. 387-410).

(Springer, 1985).

28. Oja, E. Neural networks, principal components, and subspaces. *Int. J. Neural Syst.* **1**, 61-68 (1989).

29. Xu, L. Least mean square error reconstruction principle for self-organizing neural-nets. *Neural Netw.* **6**, 627-648 (1993).

30. Chen, T., Hua, Y. & Yan, W. Y. Global convergence of Oja's subspace algorithm for principal component extraction. *IEEE Trans. Neural Netw.* **9**, 58-67 (1998).

31. Bell, A. J. & Sejnowski, T. J. An information-maximization approach to blind separation and blind deconvolution. *Neural Comput.* **7**, 1129-1159 (1995).

32. Amari, S. I., Cichocki, A. & Yang, H. H. A new learning algorithm for blind signal separation. *Adv. Neural Info. Proc. Syst.* **8**, 757-763 (1996).

33. LeCun, Y., Bottou, L., Bengio, Y. & Haffner, P. Gradient-based learning applied to document recognition. *Proc. IEEE* **86**, 2278-2324 (1998). The MNIST data set is available at http://yann.lecun.com/exdb/mnist/index.html.

34. Isomura, T. & Toyoizumi, T. On the achievability of blind source separation for high-dimensional nonlinear source mixtures. Preprint at https://arxiv.org/abs/1808.00668 (2018).

35. Dimigen, O. Optimizing the ICA-based removal of ocular EEG artifacts from free viewing

experiments. *Neuroimage* **207**, 116117 (2020).

36. Geusebroek, J. M., Burghouts, G. J. & Smeulders, A. W. The Amsterdam library of object images. *Int. J. Comput. Vis.* **61**, 103-112 (2005). The ALOI data set is available at http://aloi.science.uva.nl.

37. Yu, F., Xian, W., Chen, Y., Liu, F., Liao, M., Madhavan, V. & Darrell, T. BDD100K: A diverse driving video database with scalable annotation tooling. Preprint at https://arxiv.org/abs/1805.04687 (2018). The BDD100K data set is available at https://bdd-data.berkeley.edu.

38. Schrödinger, E. What is life? The physical aspect of the living cell and mind. (Cambridge University Press, 1944).

39. Palmer, S. E., Marre, O., Berry, M. J. & Bialek, W. Predictive information in a sensory population. *Proc. Natl. Acad. Sci. USA* **112**, 6908-6913 (2015).

40. Friston, K., Kilner, J. & Harrison, L. A free energy principle for the brain. *J. Physiol. Paris* **100**, 70-87 (2006).

41. Oymak, S., Fabian, Z., Li, M. & Soltanolkotabi, M. Generalization Guarantees for Neural Networks via Harnessing the Low-rank Structure of the Jacobian. Preprint at https://arxiv.org/abs/1906.05392 (2019).

42. Suzuki, T., Abe, H., Murata, T., Horiuchi, S., Ito, K., Wachi, T., ... & Nishimura, T. Spectral-Pruning: Compressing deep neural network via spectral analysis. Preprint at https://arxiv.org/abs/1808.08558 (2018).

43. Neftci, E. Data and power efficient intelligence with neuromorphic learning machines. *iScience* **5**, 52-68 (2018).

44. Fouda, M., Neftci, E., Eltawil, A. M. & Kurdahi, F. Independent component analysis using RRAMs. *IEEE Trans. Nanotech.* **18**, 611-615 (2018).

45. Lee, T. W., Girolami, M., Bell, A. J. & Sejnowski, T. J. A unifying information-theoretic framework for independent component analysis. *Comput. Math. Appl.* **39**, 1-21 (2000).

46. Isomura, T. & Toyoizumi, T. A local learning rule for independent component analysis. *Sci. Rep.* **6**, 28073 (2016).

47. Isomura, T. & Toyoizumi, T. Error-gated Hebbian rule: A local learning rule for principal and independent component analysis. *Sci. Rep.* **8**, 1835 (2018).

48. Dayan, P., Hinton, G. E., Neal, R. M. & Zemel, R. S. The Helmholtz machine. *Neural Comput.* **7**, 889-904 (1995).

49. Frémaux, N. & Gerstner, W. Neuromodulated spike-timing-dependent plasticity, and theory of three-factor learning rules. *Front. Neural Circuits* **9**, 85 (2016).

50. Kuśmierz, Ł., Isomura, T. & Toyoizumi, T. Learning with three factors: modulating Hebbian plasticity with errors. *Curr. Opin. Neurobiol.* **46**, 170-177 (2017).

**Acknowledgements**

We are grateful to Shun-ichi Amari for helpful discussions. This work was supported by RIKEN Center for Brain Science (T.I. and T.T.), Brain/MINDS from AMED under Grant Number JP19dm020700 (T.T.), and JSPS KAKENHI Grant Number JP18H05432 (T.T.). The funders had no role in the study design, data collection and analysis, decision to publish, or preparation of the manuscript.

**Author Contributions**

Takuya Isomura conceived and designed PredPCA, performed the mathematical analyses and simulations, and wrote the manuscript. Taro Toyoizumi supervised T.I. from the early state of this work, confirmed rigor of the mathematical analyses, and wrote the manuscript.

**Competing Interests**

The authors declare that they have no competing interests.

**Additional Information**

Correspondence and requests for materials should be addressed to T.I. or T.T.

**METHODS**

In what follows, we mathematically express the benefits of PredPCA. Methods A and B formally define the system and PredPCA. Methods C and D prove that PredPCA inherits preferable properties of both the standard PCA and AR models, and outperforms naïve PCA and AR models in terms of robustness to noise and generalization of prediction. Methods E and F demonstrate that PredPCA identifies the optimal hidden state estimator and the true system parameters of a class of canonical systems with a global convergence guarantee, owing to the asymptotic property of linear neural networks with high-dimensional inputs [34]. Supplementary Methods 1 and 2 provide the simulation protocols.

**A. System**

We suppose that a system in the external milieu is expressed as $x_{t+1} = f_t + z_t$ and $s_t = g_t + \omega_t$, where $f_t \equiv f(x_t, x_{t-1}, \ldots)$ and $g_t \equiv g(x_t)$ are nonlinear functions of $x_t$, while $z_t$ and $\omega_t$ are mutually independent white noises characterized with zero means and covariances $\Sigma_z$ and $\Sigma_\omega$. We assume that the system is in a steady state. To generate predictable dynamics, $\Sigma_z$ is assumed to be smaller than $\Sigma_x$ in magnitude; whereas, we typically consider a large $\Sigma_\omega$. Without loss of generality, we can suppose that the steady state of $x_t$ follows a distribution with zero mean and the identity covariance $\Sigma_x \equiv I$. For analysis, we consider a family of functions $f_t \equiv B\psi_t$ and $g_t \equiv A\psi_t$ spanned by nonlinear basis functions $\psi_t \equiv \psi(x_t) \in \mathbb{R}^{N_\psi}$, where $N_\psi$ denotes the number of linearly independent bases, $B \in \mathbb{R}^{N_x \times N_\psi}$ is a full-row-rank transition matrix, and $A \in \mathbb{R}^{N_s \times N_\psi}$ is a full-column-rank mapping matrix from the bases to the sensory input. Thus, equation (1) becomes

$$s_t = A\psi_t + \omega_t \qquad (8)$$

and equation (2) becomes

$$x_{t+1} = B\psi_t + z_t. \tag{9}$$

As the dimensionality of bases increase, each element of $f(x_t)$ and $g(x_t)$ asymptotically expresses an arbitrary nonlinear mapping if $A$ and $B$ are suitably selected (c.f., universality). We assume $N_x \leq N_\psi \leq N_s$ such that the system dynamics are produced by hidden states that are lower-dimensional than the observations. Although this paper supposes $\psi_t = \psi(x_t)$, the same analysis can be applied to a system comprising $\psi_t = \psi(x_t, x_{t-1}, \ldots)$ by redefining $(x_t, x_{t-1}, \ldots)$ and $(s_t, s_{t-1}, \ldots)$ as new $x_t$ and $s_t$, respectively. **Table 1** presents the glossary of expressions.

**B. Derivation of PredPCA**

PredPCA aims to minimize the multistep prediction error for predicting a 1 to $K_f$-step future of the aforementioned system by optimizing synaptic weight matrices using and only using the current and past observations $s_t, s_{t-1}, \ldots, s_{t-K_p+1}$, where $K_f$ and $K_p$ are imposed by the problem setup. Hidden states and bases $(x_t, \psi_t)$, system parameters $(A, B, \Sigma_x, \Sigma_\psi, \Sigma_z, \Sigma_\omega)$, and the numbers of hidden state and basis dimensions $(N_x, N_\psi)$ are unknown to a learner.

The error for predicting the *k*-step future is defined by $\varepsilon_{t+k|t} \equiv s_{t+k} - W^T V_k \phi_t$, where $\phi_t \equiv (s_t^T, s_{t-1}^T, \ldots, s_{t-K_p+1}^T)^T \in \mathbb{R}^{N_\phi}$ is a vector of observations, $W \in \mathbb{R}^{N_u \times N_s}$ is the transpose of the decoding synaptic weight matrix, and $V_k \in \mathbb{R}^{N_u \times N_\phi}$ is the *k*-th encoding synaptic weight matrix. Although general nonlinear bases can be used as $\phi_t$, a simple vector of observations serves the purpose of this paper. We will show below that the prediction and system identification using these linear bases are accurate when the dimensionality of inputs are sufficiently large. Minimizing $\varepsilon_{t+k|t}$ can be viewed as a generalization of the standard PCA [29] that minimizes the reconstruction error of the current observation (i.e., $\varepsilon_t^{PCA} \equiv s_t - W^T W s_t$).

Formally, the cost function of PredPCA for multistep predictions is defined by

$$L \equiv \frac{1}{2} \sum_{k=1}^{K_f} \left\langle |\varepsilon_{t+k|t}|^2 \right\rangle_q, \tag{10}$$

where $\langle \cdot \rangle_q \equiv \frac{1}{T} \sum_{t=1}^{T} \cdot$ is the expectation over the empirical distribution $q$. Solving the fixed point of the above cost function $L$ with respect to $V_k$ yields the optimal estimator. From

$$\frac{\partial L}{\partial V_k} = -W \langle \varepsilon_{t+k|t} \phi_t^T \rangle_q = -W \langle (s_{t+k} - W^T V_k \phi_t) \phi_t^T \rangle_q = 0, \tag{11}$$

under an assumption of $WW^T = I$ (which is preserved by equation (13) below), the optimal $V_k$ is found as

$$V_k = W \langle s_{t+k} \phi_t^T \rangle_q \langle \phi_t \phi_t^T \rangle_q^{-1}. \tag{12}$$

We define the maximum likelihood estimator of $s_{t+k}$ by $\mathbf{s}_{t+k|t} \equiv \mathbf{Q}_k \phi_t$, where $\mathbf{Q}_k \equiv \langle s_{t+k} \phi_t^T \rangle_q \langle \phi_t \phi_t^T \rangle_q^{-1}$ is the optimal (maximum likelihood) matrix estimator. Throughout the manuscript, a bold case variable (e.g., $\mathbf{s}_{t+k|t}$) indicates the estimator of the corresponding italic case variable (e.g., $s_{t+k}$). The k-th encoder $u_{t+k|t}$ is thus defined by $u_{t+k|t} \equiv W\mathbf{s}_{t+k|t}$. The optimal W is determined by the gradient descent on L:

$$\dot{W} \propto -\frac{\partial L}{\partial W} = \sum_{k=1}^{K_f} \left\langle u_{t+k|t}\left(s_{t+k} - W^T u_{t+k|t}\right)^T \right\rangle_q. \tag{13}$$

Equation (13) is similar to Oja's subspace rule for PCA [28] except that $\mathbf{s}_{t+k|t}$ is used instead of $s_{t+k}$ to define $u_{t+k|t}$. In this sense, PredPCA conducts post-hoc dimensionality reduction (PCA) of the predicted input. The update by equation (13) maintains $W$ as an orthogonal matrix (i.e., $WW^T = I$) throughout the learning.

The above PredPCA solution can also be obtained by eigenvalue decomposition. When $WW^T = I$, the cost function is transformed as $L = \frac{1}{2}\sum_{k=1}^{K_f} \left\langle |s_{t+k} - W^T W \mathbf{s}_{t+k|t}|^2 \right\rangle_q = \frac{K_f}{2}\left(\text{tr}[\Sigma_s] - \text{tr}[W\Sigma_\mathbf{s}^{\text{Pred}} W^T]\right)$, where $\Sigma_s \equiv \langle s_t s_t^T \rangle_q$ and $\Sigma_\mathbf{s}^{\text{Pred}} \equiv \frac{1}{K_f}\sum_{k=1}^{K_f}\langle \mathbf{s}_{t+k|t} \mathbf{s}_{t+k|t}^T \rangle_q$ are the actual and predicted input covariances calculated based on the empirical distribution, respectively. Thus, the minimization of $L$ is achieved by maximizing the second term under the constraint of $WW^T = I$ (note that this constraint is automatically satisfied by minimizing $L$). Hence, the optimal $W$ is provided as the transpose of the major eigenvectors of $\Sigma_\mathbf{s}^{\text{Pred}}$. This solution is unique up to the multiplication of an $N_u \times N_u$ orthogonal matrix from the left. The global convergence and absence of spurious solutions are guaranteed even when $W$ is computed by equation (13) because of the global convergence property of Oja's subspace rule for PCA [30]. In short, PredPCA is a convex optimization and thus it can reliably identify the optimal synaptic weight matrices $W$ and $V_1, \ldots, V_{K_f}$ for predictions, which provides the global minimum of the cost function $L$.

**C. PredPCA (but not PCA) filters out observation noise**

Here, we compare the components extracted using PredPCA and the standard PCA [28,29]. We show that only PredPCA can remove observation noise and accurately estimate the observation matrix $A$ as training sample size $T$ increases.

We introduce the expectation over true distribution $p\left(\phi_t, s_{t+1}, \ldots, s_{t+K_f}\right)$, denoted by $\langle \cdot \rangle \equiv \int \cdot \, p\left(\phi_t, s_{t+1}, \ldots, s_{t+K_f}\right) d\phi_t ds_{t+1} \cdots ds_{t+K_f}$. The empirical distribution approaches this true distribution in the large training sample size limit: $p\left(\phi_t, s_{t+1}, \ldots, s_{t+K_f}\right) = \plim_{T \to \infty} q(\phi_t, s_{t+1}, \ldots, s_{t+K_f})$. Throughout the manuscript, $\langle s_t \rangle = 0$, $\langle \psi_t \rangle = 0$, and $\langle x_t \rangle = 0$ are supposed for the sake of simplicity. The true covariance matrix of some variable $\xi_t$ is denoted by $\Sigma_\xi \equiv \mathrm{Cov}[\xi_t] \equiv \langle \xi_t \xi_t^T \rangle - \langle \xi_t \rangle \langle \xi_t^T \rangle$. Here, any estimator or statistic $\boldsymbol{\theta}$ under consideration, calculated based on the empirical distribution, can be decomposed into its true value $\theta$ and its generalization error $\delta\theta \equiv \boldsymbol{\theta} - \theta$, where $\delta\theta$ is in the $T^{-1/2}$ order (see Supplementary Methods 4 for the conditions and the proof). Below, we will decompose $\boldsymbol{\theta}$ into $\theta$ and $\delta\theta$ and then solve $\theta$ analytically.

The standard PCA conducts the eigenvalue decomposition of the actual input covariance, calculated based on the empirical distribution: $\boldsymbol{\Sigma}_s \equiv \langle s_t s_t^T \rangle_q$. The convergence to some unknown underlying distribution in the large-sample limit is a known property of PCA [51]. From equation (8), the covariance is decomposed as $\boldsymbol{\Sigma}_s = \Sigma_s + \mathcal{O}(T^{-1/2}) = A\Sigma_\psi A + \Sigma_\omega + \mathcal{O}(T^{-1/2})$ owing to the independence of $\psi_t$ and $\omega_t$. As the observation noise covariance $\Sigma_\omega$ is involved in $\boldsymbol{\Sigma}_s$, the major eigenvectors of $\boldsymbol{\Sigma}_s$ that PCA extracts are biased toward the directions of the noise's major eigenvectors. This bias is a common issue of autoencoding approaches [10,11] that renders the identification of the true system parameters difficult.

In contrast to the standard PCA, PredPCA conducts the eigenvalue decomposition of the predicted input covariance: $\mathbf{\Sigma}_\mathbf{s}^{\text{Pred}} \equiv \frac{1}{K_f}\sum_{k=1}^{K_f}\langle \mathbf{s}_{t+k|t}\mathbf{s}_{t+k|t}^T \rangle_q$. Owing to this construction, the identification of system parameters $(A, B, \Sigma_x, \Sigma_\psi, \Sigma_\omega, \Sigma_z)$ based on PredPCA is not biased by the observation noise. From the independence between $\omega_{t+k}$ and $\phi_t$, $\mathbf{s}_{t+k|t} = A\langle \psi_{t+k}\phi_t^T\rangle\Sigma_\phi^{-1}\phi_t + \mathcal{O}(T^{-1/2})$ holds. Thus, we obtain

$$\mathbf{\Sigma}_\mathbf{s}^{\text{Pred}} = A\Sigma_\psi^{\text{Pred}}A^T + \mathcal{O}\left(T^{-\frac{1}{2}}\right), \tag{14}$$

where $\Sigma_\psi^{\text{Pred}} \equiv \frac{1}{K_f}\sum_{k=1}^{K_f}\langle \psi_{t+k}\phi_t^T\rangle\Sigma_\phi^{-1}\langle \phi_t\psi_{t+k}^T\rangle$ is the predicted hidden basis covariance, calculated based on the true distribution. Applying the eigenvalue decomposition to $\mathbf{\Sigma}_\mathbf{s}^{\text{Pred}}$ provides the set of major eigenvectors $\mathbf{P}_\mathbf{s} \equiv (\mathbf{P}_{\cdot 1}, \ldots, \mathbf{P}_{\cdot N_\psi}) \in \mathbb{R}^{N_s \times N_\psi}$ that correspond to asymptotically non-zero eigenvalues of the predicted input covariance. Because of the uniqueness of the eigenvalue decomposition, $\mathbf{P}_\mathbf{s}$ converges to matrix $A$ as the number of training samples increases—up to the multiplication of a full-rank matrix $\Omega_\psi \in \mathbb{R}^{N_\psi \times N_\psi}$ from the right-hand side. Hence, we refer to $\mathbf{P}_\mathbf{s}$ as the estimator of $A$:

$$\mathbf{A} \equiv \mathbf{P}_\mathbf{s} = P_\mathbf{s} + \mathcal{O}\left(T^{-\frac{1}{2}}\right) = A\Omega_\psi^{-1} + \mathcal{O}\left(T^{-\frac{1}{2}}\right). \tag{15}$$

Here, we introduced the inverse of $\Omega_\psi$ (instead of $\Omega_\psi$ itself) for our convenience. Note that $P_\mathbf{s}$ is the set of major eigenvectors of the generalization-error-free predicted input covariance $\Sigma_\mathbf{s}^{\text{Pred}} \equiv A\Sigma_\psi A^T$. In short, PredPCA can identify matrix $A$ with asymptotically zero error without directly observing $\psi_t$ for large $T$. Notably, the number of basis dimensions $N_\psi$ is also identifiable by counting the number of asymptotically non-zero eigenvalues of $\boldsymbol{\Sigma}_\mathbf{s}^{\text{Pred}}$, which converges to the true $N_\psi$ of canonical systems for a large training sample size (see Methods D for the formal definition of the estimator $\mathbf{N}_\psi$ using the test prediction error).

In addition, multiplying $\mathbf{P}_\mathbf{s}^T$ by the predicted input yields the predicted basis estimator:

$$\boldsymbol{\psi}_{t+k|t} \equiv \mathbf{P}_\mathbf{s}^T \mathbf{s}_{t+k|t} = \Omega_\psi \langle \psi_{t+k} \phi_t^T \rangle \Sigma_\phi^{-1} \phi_t + \mathcal{O}\left(T^{-\frac{1}{2}}\right). \tag{16}$$

The last equality holds from the orthogonality of eigenvectors, i.e., $P_\mathbf{s}^T A = P_\mathbf{s}^T P_\mathbf{s} \Omega_\psi = \Omega_\psi$, and the independence between $\omega_{t+k}$ and $\phi_t$. Indeed, $u_{t+k|t}$ with optimized synaptic weight matrices is equivalent to $\boldsymbol{\psi}_{t+k|t}$ when $N_u = N_\psi$. In short, PredPCA can provide the maximum likelihood estimator of the hidden bases without directly observing $\psi_{t+k}$—up to the multiplication of the full-rank ambiguity factor $\Omega_\psi$ from the left-hand side. This ambiguity factor is safely absorbed into the definition of $\psi_t$, without changing the system dynamics, by applying the following transformations: $\Omega_\psi \psi_t \to \psi_t$, $P_\mathbf{s} = A\Omega_\psi^{-1} \to A$, and $B\Omega_\psi^{-1} \to B$. Therefore, the estimated hidden dynamics are formally homologous to the original dynamics.

In terms of conceptual innovations of PredPCA, our analyses reveal that this scheme can identify the true hidden states, parameters, and dimensionalities of a class of canonical systems (see below). In particular, the multi-time-step bases function $\phi_t$ is an essential difference between PredPCA and related methods such as TICA [14], TAE [13], and DMD [15,52]. Empirical observations highlight the importance of filtering out observation noise to reliably perform system identification (**Fig. 2**, and **Extended Data Figs. 1 and 2**). Indeed, features extracted from TICA or DMD are expressed as complex numbers, which do not match the true hidden states. Although TAE can identify matrix *A* and the extracted features are denoted in real numbers, it still fails to identify true hidden states and other parameters because it fails to filter out large observation noise (**Fig. 2b** and **Extended Data Fig. 2a**).

Furthermore, we presented an algorithm to update synaptic weights (equation (5)), which makes it easier to design a computational architecture for PredPCA. As discussed above, it is fairly straightforward to implement PredPCA in neuromorphic hardware through a previously developed local learning algorithm [46,47,53]—wherein a previous work has implemented the local algorithm using resistive random-access memories [44]. It should be emphasized that PredPCA is suitable for neuromorphic hardware relative to TCIA, TAE, and DMD because the computations for inverse matrices, complex numbers, and eigenvalue decomposition of non-symmetric matrices are intractable in neural networks. For further discussion, please refer to Supplementary Discussion.

**D. PredPCA (but not AR models) minimizes test prediction error**

A learner needs to predict the future consequences of unseen input data based on learning with a limited number of training samples. Here, we analytically solve the expectation of the PredPCA's test prediction error as a function of the training samples ($T$), encoding dimensions ($N_s$), and number of past observations used for prediction ($N_\phi = K_p N_s$). Its minimization enables a learner to maximize the generalization ability by optimizing free parameters in the network without knowing the true distribution that generates test samples.

PredPCA's test prediction error is defined as the squared error over the true distribution $p$. Meanwhile, the learning is based on the empirical distribution $q$. Thus, the test error is given as a functional of $q$:

$$L_{test}[q] \equiv \frac{1}{2}\sum_{k=1}^{K_f} \left\langle |\varepsilon_{t+k|t}[q]|^2 \right\rangle. \tag{17}$$

Here, the prediction error (which is also a functional of $q$) is given as $\varepsilon_{t+k|t}[q] \equiv s_{t+k} - \mathbf{P_s P_s^T} s_{t+k|t}$ using the major eigenvectors of the predicted input covariance $\mathbf{P_s} \equiv (\mathbf{P_{\cdot 1}}, \ldots, \mathbf{P_{\cdot N_u}}) \in \mathbb{R}^{N_s \times N_u}$ and the maximum likelihood estimator $s_{t+k|t} = \langle s_{t+k}\phi_t^T\rangle_q \langle \phi_t\phi_t^T\rangle_q^{-1} \phi_t$ computed based on the empirical distribution $q$. The generalization error of major eigenvectors $\mathbf{P_s}$ is negligible up to the leading order (see Supplementary Methods 5 for details). The $q$-dependent factor in $s_{t+k|t}$ is computed as $\langle s_{t+k}\phi_t^T\rangle_q \langle \phi_t\phi_t^T\rangle_q^{-1} = (\langle s_{t+k}\phi_t^T\rangle + \delta\langle s_{t+k}\phi_t^T\rangle_q)(\Sigma_\phi + \delta\langle \phi_t\phi_t^T\rangle_q)^{-1} = Q_k + \delta\langle(s_{t+k} - Q_k\phi_t)\phi_t^T\rangle_q \Sigma_\phi^{-1}$ up to the leading order, using the optimal mapping $Q_k \equiv \langle s_{t+k}\phi_t^T\rangle \Sigma_\phi^{-1}$ (note that $\delta\langle \cdot \rangle_q \equiv \langle \cdot \rangle_q - \langle \cdot \rangle$). Thus, the prediction error becomes $\varepsilon_{t+k|t}[q] = s_{t+k} - P_s P_s^T Q_k \phi_t - P_s P_s^T \delta\langle(s_{t+k} - Q_k\phi_t)\phi_t^T\rangle_q \Sigma_\phi^{-1} \phi_t$, where $P_s \equiv (P_{\cdot 1}, \ldots, P_{\cdot N_u}) \in \mathbb{R}^{N_s \times N_u}$ denotes the major eigenvectors of the generalization-error-free predicted input covariance $\Sigma_s^{\text{Pred}}$. Then, we define the expectation of $L_{test}[q]$ over different empirical distributions $q$, given by

$$\mathcal{L} \equiv \mathrm{E}_{\{q\}}[L_{test}[q]]. \tag{18}$$

Here, $\mathrm{E}_{\{q\}}[\cdot]$ means the expectation over different empirical distributions. The expectation over different $q$ is equivalent to the expectation over $p$ for a linear term that involves a single $\delta\langle\cdot\rangle_q$ factor. Hence, $\mathrm{E}_{\{q\}}[\delta\langle(s_{t+k} - Q_k\phi_t)\phi_t^T\rangle_q] = 0$. In contrast, a term that comprises the square of $\delta\langle\cdot\rangle_q$ yields the positive variance through the interaction of the two factors, which is computed as $\mathrm{E}_{\{q\}}[\delta\langle(s_{t+k} - Q_k\phi_t)\phi_t^T\rangle_q \Sigma_\phi^{-1} \delta\langle(s_{t+k} - Q_k\phi_t)\phi_t^T\rangle_q^T] = \frac{N_\phi}{T}(\Sigma_s - Q_k\Sigma_\phi Q_k^T)$. Therefore, we find

$$\underbrace{\mathcal{L}}_{\substack{\text{test error}\\\text{expectation}}} = \underbrace{\frac{K_f}{2}\left(\mathrm{tr}[\Sigma_s] - \mathrm{tr}[P_\mathbf{s}^T \Sigma_\mathbf{s}^{\mathrm{Pred}} P_\mathbf{s}]\right)}_{\text{training error}} + \underbrace{\frac{K_f N_\phi}{2T}\mathrm{tr}[P_\mathbf{s}^T(\Sigma_s - \Sigma_\mathbf{s}^{\mathrm{Pred}})P_\mathbf{s}]}_{\text{generalization error}} + \mathcal{O}\left(T^{-\frac{3}{2}}\right). \quad (19)$$

See Supplementary Methods 5 for the detailed derivation. This is viewed as a variant of AIC [19] and NIC [20]. For practical use, covariances and eigenvectors in equation (19) are replaced with their estimators: $\Sigma_s \to \mathbf{\Sigma}_s$, $\Sigma_\mathbf{s}^{\mathrm{Pred}} \to \mathbf{\Sigma}_\mathbf{s}^{\mathrm{Pred}}$, and $P_\mathbf{s} \to \mathbf{P}_\mathbf{s}$, where $\mathcal{L}$ does not change by these replacements in the leading order. Because $\mathrm{tr}[P_\mathbf{s}^T(\Sigma_s - \Sigma_\mathbf{s}^{\mathrm{Pred}})P_\mathbf{s}] > 0$, the generalization error monotonically increases with the dimensionality of the encoders $N_u$. Meanwhile, the reduction of the training prediction error becomes small as $N_u$ increases, and it reaches zero for $N_u > N_\psi$ due to zero eigenvalues of $\Sigma_\mathbf{s}^{\mathrm{Pred}}$. Hence, the optimal $N_u$ that minimizes $\mathcal{L}$ is less than $N_s$.

The optimal encoding dimensionality that minimizes $\mathcal{L}$ is comparable to the effective dimensionality of true hidden basis dynamics of canonical systems for large $T$. Thus, $\mathbf{N}_\psi \equiv \mathrm{argmin}_{N_u}\mathcal{L}$ provides the estimator of the true hidden basis dimensionality. In particular, $\mathbf{N}_\psi = N_\psi$ holds when $T$ is larger than a large finite constant $T_\psi^c \equiv N_\phi \mathrm{tr}[\Sigma_s - \Sigma_\mathbf{s}^{\mathrm{Pred}}]/(\Lambda_\mathbf{s})_{N_\psi N_\psi}$, where $(\Lambda_\mathbf{s})_{N_\psi N_\psi}$ is the $N_\psi$-th (i.e., the smallest non-zero) eigenvalue of $\Sigma_\mathbf{s}^{\mathrm{Pred}}$. In contrast, equation (19) with $N_u = N_s$ provides the test prediction error of an AR model that does not consider hidden states: $\mathcal{L}_{AR} = \frac{K_f}{2}\left(1 + \frac{N_\phi}{T}\right)\mathrm{tr}[\Sigma_s - \Sigma_\mathbf{s}^{\mathrm{Pred}}]$. As some components of $\Sigma_\omega$ are generally perpendicular to $P_\mathbf{s}$, $\mathrm{tr}[P_\mathbf{s}^T(\Sigma_s - \Sigma_\mathbf{s}^{\mathrm{Pred}})P_\mathbf{s}] < \mathrm{tr}[\Sigma_s - \Sigma_\mathbf{s}^{\mathrm{Pred}}]$ for $N_u < N_s$. This means that the test prediction error of PredPCA with optimal $N_u$ is smaller than that of AR models. Hereafter, we suppose $N_u = \mathbf{N}_\psi = N_\psi$.

**E. Asymptotic linearization theorem guarantees PredPCA to identify true hidden states with an accuracy guarantee**

The asymptotic linearization theorem [34] was originally introduced to guarantee accurate extraction of independently and identically distributed hidden sources from its high-dimensional nonlinear transformations. In this paper, we use this theorem to prove that the true hidden state $x_t \in \mathbb{R}^{N_x}$ is accurately estimated from its unknown nonlinear transformation $\psi(x_t) \in \mathbb{R}^{N_\psi}$ with asymptotically zero element-wise error as $N_x$ and $N_\psi/N_x$ (and $T$) diverge. In this section, we suppose that $\psi(x_t)$ is expressed in a specific but generic form of two-layered structure, $\psi(x_t) = C\rho(Rx_t + r)$. Here, the elements of $R \in \mathbb{R}^{N_\psi \times N_x}$ and $r \in \mathbb{R}^{N_\psi}$ are fixed Gaussian random variables independently drawn from $\mathcal{N}[0, 1/N_x]$; $C \in \mathbb{R}^{N_\psi \times N_\psi}$ is a matrix whose elements are, on average, in the order of $N_\psi^{-1/2}$; and $\rho(\cdot): \mathbb{R} \mapsto \mathbb{R}$ is an odd nonlinear function, where the correlation between $\rho(\xi)$ and unit Gaussian variable $\xi \sim \mathcal{N}[0,1]$ is not close to zero. When $N_\psi$ is large, each element of $\psi(x_t)$ can represent an arbitrary nonlinear mapping of $x_t$ by adjusting $C$ (c.f., universality) [54–57].

The assumption behind the theorem is as follows: (1) the elements of hidden states $x_t$ are not strongly dependent on each other (where zero mean and identity covariance matrix are supposed without loss of generality), in the sense that the average of higher-order correlations of $x_t$'s elements asymptotically vanish for large $N_x$ with less than the order of 1; and that (2) the matrix components of $C$ that are parallel to $R$ are not too small compared to the other components (i.e., the mapping is not very close to a singular mapping)—namely, the ratio of the minimum eigenvalue of $R^T C^T C R$ to the maximum eigenvalue of $CC^T$ is assumed to be much greater than 1. Note that $R^T R = \mathcal{O}(N_\psi / N_x)$ is much greater than 1, so the condition (2) is easily satisfied when singular values of $C$ are of order 1. The asymptotic linearization theorem states that under these two conditions, covariance $\Sigma_\psi$ has a clear spectrum gap that separates major and minor components, where the major and minor components correspond to linear and nonlinear transformations of the true hidden states, respectively.

Let $P \in \mathbb{R}^{N_\psi \times N_x}$ be the set of the first to $N_x$-th major eigenvectors of $\Sigma_\psi$, and $\Lambda \in \mathbb{R}^{N_x \times N_x}$ be the diagonal matrix of the corresponding eigenvalues. The asymptotic linearization theorem proved that applying PCA to $\psi(x_t)$ provides accurate estimation of $x_t$ up to the multiplication of a fixed orthogonal matrix $\Omega$; i.e., $\Lambda^{-1/2} P^T \psi(x_t) = \Omega x_t + \mathcal{O}(\sigma_x)$. Here, $\sigma_x = \sqrt{\left(\overline{\rho^2}/\overline{\rho'}^2 - 1\right)(1+\lambda)N_x/N_\psi + \overline{\rho'''}^2/\left(2\overline{\rho'}^2 N_x\right)}$ is the standard deviation of the linearization error, where $\overline{\rho^2} \equiv \int \rho^2(\xi) p(\xi) d\xi$, $\overline{\rho'} \equiv \int \frac{d\rho(\xi)}{d\xi} p(\xi) d\xi$, and $\overline{\rho'''} \equiv \int \frac{d^3 \rho(\xi)}{d\xi^3} p(\xi) d\xi$ are statistics of the nonlinear function $\rho$ over unit Gaussian variable $\xi$, and $\lambda$ is an order-one constant determined by the characteristics of $C$. The linearization error monotonically decreases as the system size increases (i.e., when $N_\psi/N_x$ and $N_x$ diverge)—asymptotically achieving the zero-element-wise-error hidden state estimation in the large system size limit. Please refer to [34] for further details.

This theorem can be applied to the estimator of $\psi(x_t)$. Let $\mathbf{P}_\psi \in \mathbb{R}^{N_\psi \times N_x}$ be the major eigenvectors of $\mathbf{\Sigma}_\psi$ (see equation (23) below for its definition and analytical solution), and $\mathbf{\Lambda}_\psi \in \mathbb{R}^{N_x \times N_x}$ be the corresponding eigenvalues. The hidden state estimator is given as

$$\mathbf{x}_{t+k|t} \equiv \Lambda_\psi^{-\frac{1}{2}} P_\psi^T \mathbf{\psi}_{t+k|t} = \Lambda_\psi^{-\frac{1}{2}} P_\psi^T \Omega_\psi \langle \psi_{t+k} \phi_t \rangle \Sigma_\phi^{-1} \phi_t + \mathcal{O}\left(T^{-\frac{1}{2}}\right). \tag{20}$$

From the asymptotic linearization theorem, $\Lambda_\psi^{-1/2} P_\psi^T \Omega_\psi \psi_{t+k} = \Omega_x x_{t+k} + \mathcal{O}(\sigma_x)$ holds, where $\Omega_x \in \mathbb{R}^{N_x \times N_x}$ is a fixed orthogonal matrix. Here, we treated $\Omega_\psi \psi_{t+k}$ as a nonlinear function of $x_{t+k}$ and applied the theorem. Thus, equation (20) is solved analytically as

$$\mathbf{x}_{t+k|t} = \Omega_x \langle x_{t+k} \phi_t \rangle \Sigma_\phi^{-1} \phi_t + \mathcal{O}\left(T^{-\frac{1}{2}}\right) + \mathcal{O}(\sigma_x). \tag{21}$$

This result shows that the maximum likelihood estimator of $x_{t+k}$ based on $\phi_t$, $\langle x_{t+k} \phi_t \rangle \Sigma_\phi^{-1} \phi_t$, is available (up to the ambiguity factor $\Omega_x$, and the order $T^{-1/2}$ and $\sigma_x$ small error terms), despite the fact that PredPCA is unsupervised learning that does not use any explicit data of $x_{t+k}$ for training. Similar to $\Omega_\psi$, the ambiguity of $\Omega_x$ can be absorbed into the definition of $x_t$, without changing any system dynamics, by applying the following transformations: $\Omega_x x_t \to x_t$, $\Omega_x B \to B$, $\Omega_x z_t \to z_t$, and $R \Omega_x^{-1} \to R$. Notably, the number of state dimensions $N_x$ is also identifiable by defining the estimator $\mathbf{N}_x$ as the largest spectrum gap of $\mathbf{\Sigma}_\psi$, which is guaranteed to converge to true $N_x$ when $\sigma_x$ is smaller than a small positive constant $\sigma_x^c$ and $T$ is larger than a large finite constant $T_x^c$.

It is well-known that conventional nonlinear blind source separation (BSS) approaches using nonlinear neural networks (e.g., nonlinear ICA) do not guarantee the identification of true hidden sources under the general nonlinear BSS setup [58,59]. In contrast, it is remarkable that the asymptotic linearization theorem mathematically guarantees the achievability of the nonlinear BSS when $N_\psi \gg N_x \gg 1$ [34].

**F. PredPCA identifies true parameters of canonical systems with an accuracy guarantee**

We demonstrated above that PredPCA can identify the true observation matrix $A$. Here, we show that it can also identify other system parameters $B$, $\Sigma_\psi$, $\Sigma_x$, $\Sigma_\omega$, and $\Sigma_z$ asymptotically—if the assumptions of the asymptotic linearization theorem are met and the number of training samples is large.

These parameter identifications are based on the linearized transition mapping from $\psi_t$ to $\psi_{t+1}$, denoted by $\Psi$; thus, we first compute the estimator of $\Psi$. We decompose $\psi_{t+1}$ as $\psi_{t+1} = \Psi \psi_t + \Delta \psi_{t+1|t} + \mathcal{O}(z_t)$, where $\Psi = \langle \psi_{t+1} \psi_t^T \rangle \Sigma_\psi^{-1}$ is the optimal basis transition matrix, $\Delta \psi_{t+1|t}$ is the linearization error that is perpendicular to both $\psi_t$ and $z_t$, and $\mathcal{O}(z_t)$ is a term related to small noise $z_t$. This $\Psi$ can be viewed as a finite basis size version of the Koopman operator [52,60]. The basis estimator based on the current input is defined as $\boldsymbol{\psi}_{t|t} \equiv \mathbf{P}_\mathbf{s}^T s_t$ and computed as $\boldsymbol{\psi}_{t|t} = \Omega_\psi \psi_t + P_\mathbf{s}^T \omega_t + \mathcal{O}(T^{-1/2})$. Using this, we have $\langle \boldsymbol{\psi}_{t+k|t+k} \boldsymbol{\psi}_{t|t}^T \rangle_q = \langle (\Omega_\psi \psi_{t+k} + P_\mathbf{s}^T \omega_{t+k})(\Omega_\psi \psi_t + P_\mathbf{s}^T \omega_t)^T \rangle + \mathcal{O}(T^{-1/2}) = \Omega_\psi \langle \psi_{t+k} \psi_t^T \rangle \Omega_\psi^T + \mathcal{O}(T^{-1/2})$ as the observation noise is white and independent of $\psi_t$ and $\psi_{t+k}$. In particular, $\langle \psi_{t+1} \psi_t^T \rangle = \Psi \Sigma_\psi$ and $\langle \psi_{t+2} \psi_t^T \rangle = \Psi^2 \Sigma_\psi + \langle \Delta\psi_{t+2|t+1} \psi_t^T \rangle$ hold. Thus, we obtain the following estimator of the basis transition matrix:

$$\boldsymbol{\Psi} \equiv \langle \boldsymbol{\psi}_{t+2|t+2} \boldsymbol{\psi}_{t|t}^T \rangle_q \langle \boldsymbol{\psi}_{t+1|t+1} \boldsymbol{\psi}_{t|t}^T \rangle_q^{-1} = \Omega_\psi \Psi \Omega_\psi^{-1} + \Omega_\psi \langle \Delta\psi_{t+2|t+1} \psi_t^T \rangle \Sigma_\psi^{-1} \Psi^{-1} \Omega_\psi^{-1} + \mathcal{O}\left(T^{-\frac{1}{2}}\right)$$

$$= \Omega_\psi \Psi \Omega_\psi^{-1} + \mathcal{O}\left(T^{-\frac{1}{2}}\right) + \mathcal{O}(\sigma_\psi). \quad (22)$$

This estimator converges to the optimal $\Psi$ up to the ambiguity of $\Omega_\psi$ for large $T$ and $N_\psi$. The variance of the linearization error term $\mathcal{O}(\sigma_\psi)$ is in the same order as the variance of nonlinearly transformed components of $x_t$ that are involved in $\psi_t$; thus, using the asymptotic linearization theorem [34], we compute the variance of the nonlinear components and obtain $\sigma_\psi = \sqrt{\left(\overline{\rho^2} - \overline{\rho'}^2\right)/N_\psi}$ as the order (see Supplementary Methods 6 for further details).

Next, we compute the covariance matrices of hidden bases and observation noise. By multiplying the inverse of $\Psi$ with $\langle \boldsymbol{\psi}_{t+1|t+1} \boldsymbol{\psi}_{t|t}^T \rangle_q = \Omega_\psi \Psi \Sigma_\psi \Omega_\psi^T + \mathcal{O}(T^{-1/2})$, we find the hidden basis covariance estimator (symmetrized version) as

$$\boldsymbol{\Sigma}_\psi \equiv \frac{1}{2}\left(\boldsymbol{\Psi}^{-1} \langle \boldsymbol{\psi}_{t+1|t+1} \boldsymbol{\psi}_{t|t}^T \rangle_q + \langle \boldsymbol{\psi}_{t|t} \boldsymbol{\psi}_{t+1|t+1}^T \rangle_q \boldsymbol{\Psi}^{-T}\right) = \Omega_\psi \Sigma_\psi \Omega_\psi^T + \mathcal{O}\left(T^{-\frac{1}{2}}\right) + \mathcal{O}(\sigma_\psi). \quad (23)$$

See Supplementary Methods 6 for the order of the linearization error term. Using this $\boldsymbol{\Sigma}_\psi$, the observation noise covariance estimator is given as

$$\boldsymbol{\Sigma}_\omega \equiv \boldsymbol{\Sigma}_s - \mathbf{A}\boldsymbol{\Sigma}_\psi \mathbf{A}^T = \Sigma_s - A\Sigma_\psi A^T + \mathcal{O}\left(T^{-\frac{1}{2}}\right) + \mathcal{O}(\sigma_\psi) = \Sigma_\omega + \mathcal{O}\left(T^{-\frac{1}{2}}\right) + \mathcal{O}(\sigma_\psi). \quad (24)$$

Finally, we estimate the state transition matrix and covariance matrices of hidden states and process noise. From equation (9) and the independence between $z_{t+2}$ and $\phi_t$, $\langle x_{t+2} \phi_t^T \rangle =$ $B\langle\psi_{t+1}\phi_t^T\rangle$ holds. Thus, equation (21) for k = 2 becomes $\mathbf{x}_{t+2|t} = (\Omega_x B\Omega_\psi^{-1})\Omega_\psi\langle\psi_{t+1}\phi_t\rangle\Sigma_\phi^{-1}\phi_t + \mathcal{O}(T^{-1/2}) + \mathcal{O}(\sigma_x)$. Hence, using equation (16), we find the following estimator of the transition matrix:

$$\mathbf{B} \equiv \langle\mathbf{x}_{t+2|t}\mathbf{\Psi}_{t+1|t}^T\rangle_q \langle\mathbf{\Psi}_{t+1|t}\mathbf{\Psi}_{t+1|t}^T\rangle_q^{-1} = \Omega_x B\Omega_\psi^{-1} + \mathcal{O}\left(T^{-\frac{1}{2}}\right) + \mathcal{O}(\sigma_x). \tag{25}$$

The hidden state covariance estimator is given by $\mathbf{\Sigma}_x \equiv \Sigma_x \equiv I$ as we defined $\Sigma_x$ so. Thus, as equation (9) yields $\Sigma_x = B\Sigma_\psi B^T + \Sigma_z$, the process noise covariance estimator is given by

$$\mathbf{\Sigma}_z \equiv \mathbf{\Sigma}_x - \mathbf{B}\mathbf{\Sigma}_\psi \mathbf{B}^T = \Omega_x \Sigma_z \Omega_x^T + \mathcal{O}\left(T^{-\frac{1}{2}}\right) + \mathcal{O}(\sigma_x). \tag{26}$$

In summary, PredPCA could identify the true system parameters $A$, $B$, $\Psi$, $\Sigma_\psi$, $\Sigma_x$, $\Sigma_\omega$, and $\Sigma_z$ with a global convergence guarantee as the system and training sample sizes increase, using noisy observations only—up to the full-rank linear transformations $(\Omega_\psi, \Omega_x)$ that do not change the system dynamics. When $z_t$ and $\omega_t$ are white Gaussian noises, these parameters are sufficient to identify (i.e., uniquely determine) the canonical system. The aforementioned analyses hold true even when $z_t$ and $\omega_t$ are white non-Gaussian noises, although in this case, unsupervised identification of the third or higher order moments of $z_t$ and $\omega_t$ has not yet been established. The zero-element-wise-error identification of these parameters will be asymptotically achieved when $N_\psi/N_x$, $N_x$, and $T$ diverge. This global convergence guarantee is an advantage of PredPCA compared with conventional system identification approaches [18,61]. If $\psi(x_t)$ is a linear function of $x_t$, the true system becomes a linear system and thus provides $\sigma_x = \sigma_\psi = 0$; hence, PredPCA is guaranteed to identify the true system parameters with zero error as the increasing training samples, when $N_x \leq N_s$. **Table 2** summarizes the definitions and analytical solutions of these estimators. Every estimator can be computed by following the definition, where its analytical solution and accuracy have been proven theoretically.

The identification of system properties using PredPCA was empirically demonstrated with the example of handwritten digit images (**Fig. 2**, **Extended Data Fig. 1**, and **Supplementary Fig. 1**). Although it is difficult to prove what the true generative process is for rotating 3D objects (**Fig. 3** and **Supplementary Fig. 2**) or natural scenes (**Fig. 4** and **Extended Data Fig. 5**), empirical observations suggest that PredPCA can extract features relevant to generalized predictions. At least, PredPCA was able to identify the angles of rotating objects (**Fig. 3c** and **Supplementary Fig. 2**) and lateral motion in natural scenes (**Fig. 4c** and **Extended Data Fig. 5b**), indicating the identification of a part of their generative processes. These observations imply that the outcomes of PredPCA capture the plausible properties of natural data.

**References**

51. Zhu, B., Jiao, J. & Tse, D. Deconstructing generative adversarial networks. *IEEE Trans. Inf. Theory* doi:10.1109/TIT.2020.2983698 (2020).

52. Lusch, B., Kutz, J. N. & Brunton, S. L. Deep learning for universal linear embeddings of nonlinear dynamics. *Nat. Comm.* **9**, 4950 (2018).

53. Isomura, T. & Toyoizumi, T. Multi-context blind source separation by error-gated Hebbian rule.

*Sci. Rep.* **9**, 7127 (2019).

54. Hornik, K., Stinchcombe, M. & White, H. Multilayer feedforward networks are universal approximators. *Neural Netw.* **2**, 359-366 (1989).

55. Barron, A. R. Universal approximation bounds for superpositions of a sigmoidal function. *IEEE Trans. Info. Theory* **39**, 930-945 (1993).

56. Rahimi, A. & Recht, B. Uniform approximation of functions with random bases. In *Proceedings of the 46th Annual Allerton Conference on Communication, Control, and Computing* 555-561 (2008).

57. Rahimi, A. & Recht, B. Weighted sums of random kitchen sinks: Replacing minimization with randomization in learning. *Adv. Neural Info. Process. Sys.* **21**, 1313-1320 (2008).

58. Hyvärinen, A. & Pajunen, P. Nonlinear independent component analysis: Existence and uniqueness results. *Neural Netw.* **12**, 429-439 (1999).

59. Jutten, C. & Karhunen, J. Advances in blind source separation (BSS) and independent component analysis (ICA) for nonlinear mixtures. *Int. J. Neural Syst.* **14**, 267-292 (2004).

60. Koopman, B. O. Hamiltonian systems and transformation in Hilbert space. *Proc. Natl. Acad. Sci. USA* **17**, 315-318 (1931).

61. Ljung, L. Asymptotic behavior of the extended Kalman filter as a parameter estimator for linear systems. *IEEE Trans. Automat. Contr.* **24**, 36-50 (1979).

Table 1. Glossary of expressions.

| Expression | Description |
|---|---|
| $s_t$ | Observation |
| $\psi_t$ | Hidden bases |
| $x_t$ | Hidden states |
| $\omega_t$ | Observation noise |
| $z_t$ | Process noise |
| $A$ | Observation matrix |
| $B$ | State transition matrix |
| $\Sigma_s, \Sigma_\psi, \Sigma_x, \Sigma_\omega, \Sigma_z$ | Covariance matrices of $s_t, \psi_t, x_t, \omega_t, z_t$ |
| $N_s$ | Dimensionality of observation |
| $N_\psi$ | Dimensionality of hidden bases |
| $N_x$ | Dimensionality of hidden states |
| $u_{t+k|t}$ | Encoders |
| $\phi_t$ | Basis functions |
| $V$ | Encoding synaptic weight matrix |
| $W$ | Transpose of decoding synaptic weight matrix |
| $N_u$ | Dimensionality of encoders |
| $N_\phi$ | Dimensionality of basis functions |
| $\langle \cdot \rangle_q$ | Expectation over empirical distribution $q$ |
| $\langle \cdot \rangle$ | Expectation over true distribution $p$ |

**Table 2.** Definitions and analytical solutions of estimators.

| Estimator | Definition | Analytical solution |
|---|---|---|
| $\mathbf{s}_{t+k\|t}$ | $\langle s_{t+k}\phi_t^T\rangle_q \langle \phi_t\phi_t^T\rangle_q^{-1}\phi_t$ | $\langle s_{t+k}\phi_t^T\rangle \Sigma_\phi^{-1}\phi_t + \mathcal{O}(T^{-1/2})$ |
| $\boldsymbol{\psi}_{t+k\|t}$ | $\mathbf{P}_s^T \mathbf{s}_{t+k\|t}$ | $\Omega_\psi \langle \psi_{t+k}\phi_t^T\rangle \Sigma_\phi^{-1}\phi_t + \mathcal{O}(T^{-1/2})$ |
| $\mathbf{x}_{t+k\|t}$ | $\Lambda_\psi^{-1/2}\mathbf{P}_\psi^T \boldsymbol{\psi}_{t+k\|t}$ | $\Omega_x \langle x_{t+k}\phi_t^T\rangle \Sigma_\phi^{-1}\phi_t + \mathcal{O}(T^{-1/2}) + \mathcal{O}(\sigma_x)$ |
| $\mathbf{A}$ | $\mathbf{P}_s$ | $A\Omega_\psi^{-1} + \mathcal{O}(T^{-1/2})$ |
| $\mathbf{B}$ | $\langle \mathbf{x}_{t+2\|t}\boldsymbol{\psi}_{t+1\|t}^T\rangle_q \langle \boldsymbol{\psi}_{t+1\|t}\boldsymbol{\psi}_{t+1\|t}^T\rangle_q^{-1}$ | $\Omega_x B \Omega_\psi^{-1} + \mathcal{O}(T^{-1/2}) + \mathcal{O}(\sigma_x)$ |
| $\boldsymbol{\Psi}$ | $\langle \boldsymbol{\psi}_{t+2\|t+2}\boldsymbol{\psi}_{t\|t}^T\rangle_q \langle \boldsymbol{\psi}_{t+1\|t+1}\boldsymbol{\psi}_{t\|t}^T\rangle_q^{-1}$ | $\Omega_\psi \Psi \Omega_\psi^{-1} + \mathcal{O}(T^{-1/2}) + \mathcal{O}(\sigma_\psi)$ |
| $\boldsymbol{\Sigma}_s$ | $\langle s_t s_t^T\rangle_q$ | $\Sigma_s + \mathcal{O}(T^{-1/2})$ |
| $\boldsymbol{\Sigma}_\psi$ | $\frac{1}{2}\left(\boldsymbol{\Psi}^{-1}\langle \boldsymbol{\psi}_{t+1\|t+1}\boldsymbol{\psi}_{t\|t}^T\rangle_q + \langle \boldsymbol{\psi}_{t\|t}\boldsymbol{\psi}_{t+1\|t+1}^T\rangle_q \boldsymbol{\Psi}^{-T}\right)$ | $\Omega_\psi \Sigma_\psi \Omega_\psi^T + \mathcal{O}(T^{-1/2}) + \mathcal{O}(\sigma_\psi)$ |
| $\boldsymbol{\Sigma}_x$ | $I$ | $\Sigma_x \equiv I$ |
| $\boldsymbol{\Sigma}_\omega$ | $\boldsymbol{\Sigma}_s - \mathbf{A}\boldsymbol{\Sigma}_\psi \mathbf{A}^T$ | $\Sigma_\omega + \mathcal{O}(T^{-1/2}) + \mathcal{O}(\sigma_\psi)$ |
| $\boldsymbol{\Sigma}_z$ | $\boldsymbol{\Sigma}_x - \mathbf{B}\boldsymbol{\Sigma}_\psi \mathbf{B}^T$ | $\Omega_x \Sigma_z \Omega_x^T + \mathcal{O}(T^{-1/2}) + \mathcal{O}(\sigma_x)$ |
| $\mathbf{N}_\psi$ | $\underset{N_u}{\arg\min}\,\mathcal{L}$ | Converge to $N_\psi$ when $T > T_\psi^c$ |
| $\mathbf{N}_x$ | Largest spectrum gap of $\boldsymbol{\Sigma}_\psi$ | Converge to $N_x$ when $T > T_x^c$ and $\sigma_x < \sigma_x^c$ |

$\mathbf{P}_s$ and $\mathbf{P}_\psi$ are sets of the major eigenvectors of $\boldsymbol{\Sigma}_s^{\text{Pred}}$ and $\boldsymbol{\Sigma}_\psi$, respectively; full-rank square matrix $\Omega_\psi$ and orthogonal matrix $\Omega_x$ are ambiguity factors; $\mathcal{O}(\sigma_x) = \mathcal{O}(\sqrt{N_x/N_\psi}) + \mathcal{O}(N_x^{-1/2})$ and $\mathcal{O}(\sigma_\psi) = \mathcal{O}(N_\psi^{-1/2})$ are linearization errors, where $\sigma_x = \sigma_\psi = 0$ for any linear system; $T_x^c, T_\psi^c < \infty$ are finite large constants and $\sigma_x^c > 0$ is a small positive constant.